\documentclass[sigconf, authorversion]{acmart}

\AtBeginDocument{%
  }

\copyrightyear{2024}
\acmYear{2024}
\setcopyright{acmlicensed}\acmConference[CIKM '24]{Proceedings of the 33rd ACM International Conference on Information and Knowledge Management}{October 21--25, 2024}{Boise, ID, USA}
\acmBooktitle{Proceedings of the 33rd ACM International Conference on Information and Knowledge Management (CIKM '24), October 21--25, 2024, Boise, ID, USA}
\acmDOI{10.1145/3627673.3679822}
\acmISBN{979-8-4007-0436-9/24/10}

\acmSubmissionID{fp1994}

\usepackage{graphicx} 
\usepackage{dsfont}
\usepackage{xcolor}
\usepackage{booktabs}
\usepackage{mathrsfs}
\usepackage{amsthm}
\usepackage{algorithmic}
\usepackage{algorithm}
\usepackage{subcaption}

\usepackage{color}
\usepackage{tabularray}
\usepackage{multirow}
\begin{document}

\title{GAS-Norm: Score-Driven Adaptive Normalization for
Non-Stationary Time Series Forecasting in Deep Learning}

\author{Edoardo Urettini}
\affiliation{%
  \institution{University of Pisa}
  \institution{Scuola Normale Superiore}
  \city{Pisa}
  \country{Italy}
}
\email{edoardo.urettini@sns.it}
\authornote{All authors contributed equally to the paper}

\author{Daniele Atzeni}
\affiliation{%
  \institution{University of Pisa}
  \city{Pisa}
  \country{Italy}
}
\email{daniele.atzeni@phd.unipi.it}
\authornotemark[1]

\author{Reshawn J. Ramjattan}
\affiliation{%
  \institution{University of Pisa}
  \city{Pisa}
  \country{Italy}
}
\email{reshawn.ramjattan@phd.unipi.it}
\authornotemark[1]

\author{Antonio Carta}
\affiliation{%
  \institution{University of Pisa}
  \city{Pisa}
  \country{Italy}
}
\email{antonio.carta@unipi.it}
\authornotemark[1]


\begin{abstract}
Despite their popularity, deep neural networks (DNNs) applied to time series forecasting often fail to beat simpler statistical models. One of the main causes of this suboptimal performance is the data non-stationarity present in many processes. In particular, changes in the mean and variance of the input data can disrupt the predictive capability of a DNN. In this paper, we first show how DNN forecasting models fail in simple non-stationary settings. We then introduce GAS-Norm, a novel methodology for adaptive time series normalization and forecasting based on the combination of a Generalized Autoregressive Score (GAS) model and a Deep Neural Network. The GAS approach encompasses a score-driven family of models that estimate the mean and variance at each new observation, providing updated statistics to normalize the input data of the deep model. The output of the DNN is eventually denormalized using the statistics forecasted by the GAS model,  resulting in a hybrid approach that leverages the strengths of both statistical modeling and deep learning. The adaptive normalization improves the performance of the model in non-stationary settings. The proposed approach is model-agnostic and can be applied to any DNN forecasting model. To empirically validate our proposal, we first compare GAS-Norm with other state-of-the-art normalization methods. We then combine it with state-of-the-art DNN forecasting models and test them on real-world datasets from the Monash open-access forecasting repository. Results show that deep forecasting models improve their performance in 21 out of 25 settings when combined with GAS-Norm compared to other normalization methods.
\end{abstract}



\maketitle

\section{Introduction}
Time series forecasting has played a crucial role in decision-making and planning, becoming prevalent in a variety of real-world scenarios, such as economics \cite{sezer2020financial}, health care \cite{miotto2018deep}, and energy consumption planning \cite{deb2017review}. Following the exciting results in natural language processing \cite{otter2020survey} and computer vision \cite{guo2016deep}, nowadays deep neural networks (DNNs) have been applied to forecasting problems \cite{sengupta2020review}.

However, contrary to what occurs in other domains, DNNs do not seem to excel in time series forecasting. In this area, they achieve forecasting capabilities that are often comparable to classic statistical models \cite{godahewaMonashTimeSeries2021}. Among the most accredited explanations of this behavior, researchers identified signal-to-noise ratio and non-stationarity of the input data as likely causes \cite{hussain2008financial,kim2004artificial}. Indeed, out-of-distribution data can severely impact the outputs of strongly nonlinear machine learning models, which are easily subject to overfitting and performance degradation when input data changes in scale \cite{quinonero2022dataset}. 
\begin{figure*}[t]  
    \centering
    \parbox[t]{1\textwidth}{
        \begin{subfigure}[t]{0.33\textwidth}
            \centering
            \includegraphics[width=\textwidth]{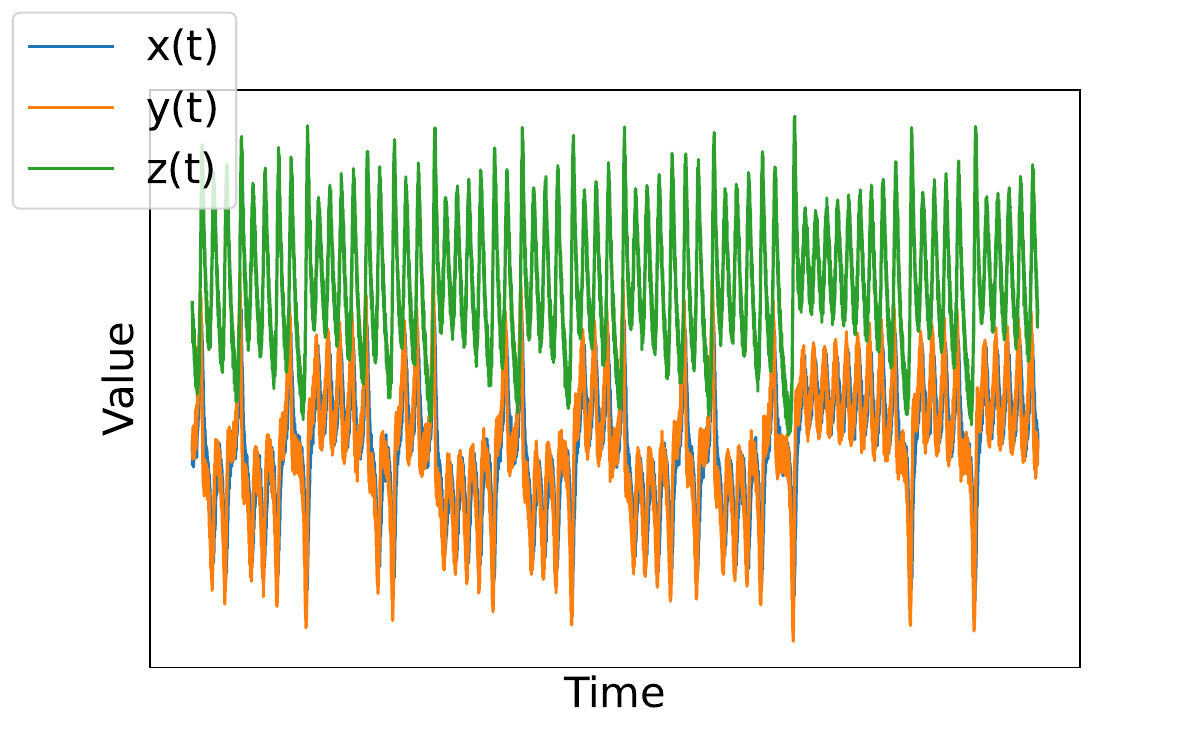}
            \caption{}
            \label{subfig:Lorenz}
        \end{subfigure}
        \hfill
        \begin{subfigure}[t]{0.33\textwidth}
            \centering
            \includegraphics[width=\textwidth]{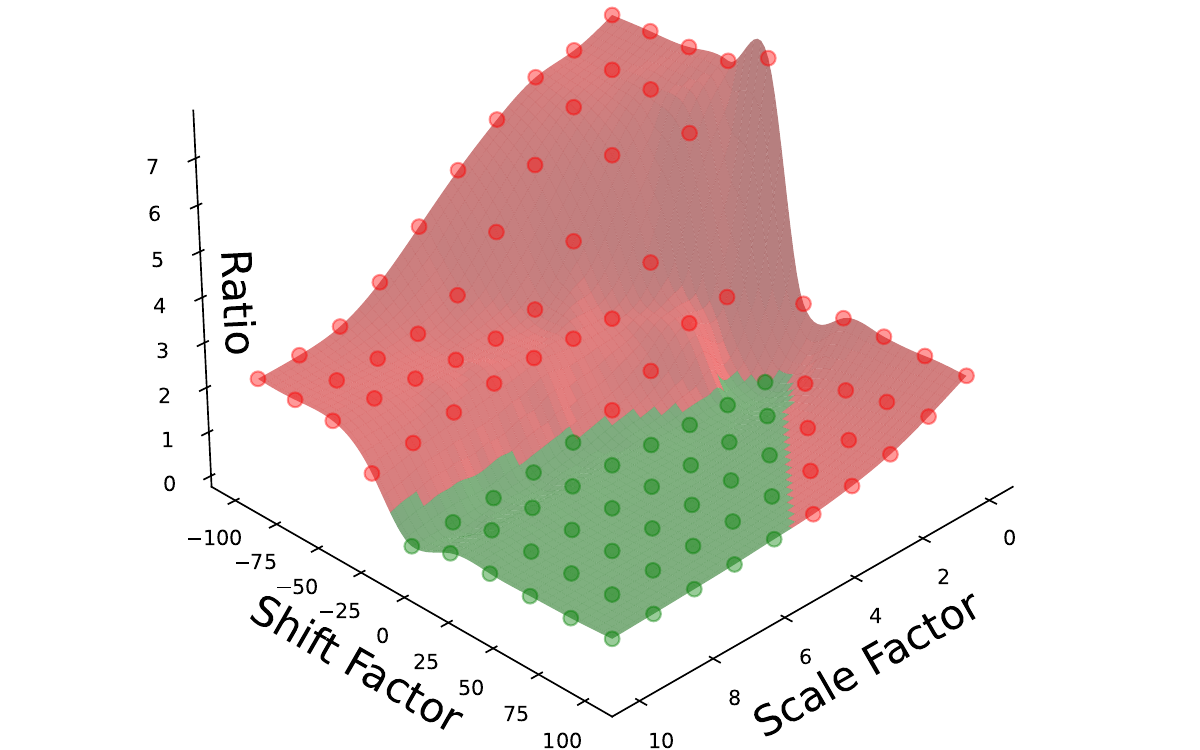}
            \caption{}
            \label{subfig:Ratio}
        \end{subfigure}
        \hfill
        \begin{subfigure}[t]{0.33\textwidth}
            \centering
            \includegraphics[width=\textwidth]{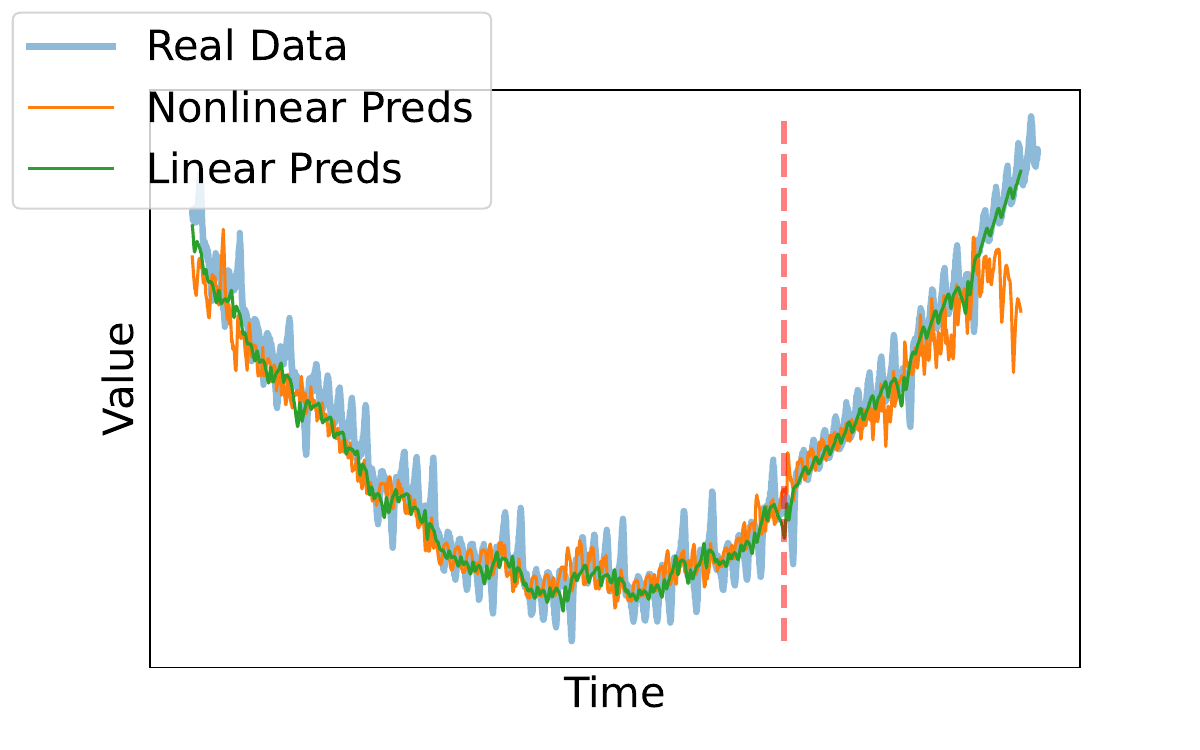}
            \caption{}
            \label{subfig:example_trend}
        \end{subfigure}
    }
    \caption{(a) The evolving coordinates of a 3D Lorenz attractor. (b) The ratio between the MSE of the nonlinear MLP and the MSE of the linear model. The green area is where the nonlinear model is better. (c)  Prediction of the 500-steps ahead value done by the linear and the nonlinear model. The vertical line shows the end of the training data.}
\end{figure*}

To avoid this, input normalization became a standard practice in deep learning. It also proved to be useful in boosting the optimization and robustness of the models by having all input features on the same scale \cite{sola1997importance}. Common estimation of input distribution parameters (e.g., mean and variance) on the training set is enough to ensure normalization if the input is stationary. However, for most time series scenarios, data usually comes from a non-stationary environment. In that case, the input distribution can change over time, requiring a different normalization approach. Given this distribution change, the normalization of the input has the additional duty of generalizing the model knowledge to the new input distribution \cite{kimReversibleInstanceNormalization2021}.

In this work, we propose GAS-Norm, a novel normalizing approach that combines DNNs and Generalized Autoregressive Score (GAS) models, a class of statistical autoregressive models developed to handle time series data with time-varying parameters \cite{crealGeneralizedAutoregressiveScore2013}. Thanks to GAS models' autoregressive nature and a training phase independent of the deep model, our normalizing module can be used in combination with a wide variety of deep forecasting architectures, whether they use autoregressive prediction or not, and with input and output time series of any length. To control the normalization strength we extend the GAS formulation with an additional parameter that allows us to set the update speed of the model. Thus, we can control how much variability is present in the data the deep model must handle during forecasting. The final output of the model combines the DNN outputs and the statistics predicted by the GAS model with a denormalization step.

We test our method against state-of-the-art dynamic normalization techniques and in combination with state-of-the-art forecasting models. The testing is done using 7 synthetic and real-world benchmarks. Using GAS-Norm improves the results of deep forecasting methods across 21 out of 25 configurations of datasets and deep learning architectures. The code for this implementation and evaluation is public and open source\footnote{\url{https://github.com/edo-urettini/GAS_Norm}\\ \url{https://github.com/daniele-atzeni/GAS-Norm-gluonTS}}. 

\section{Related Works}

A variety of deep neural network architectures have been adapted to solve forecasting problems, from recurrent autoregressive networks like DeepAR~\cite{salinasDeepARProbabilisticForecasting2020}, to convolutional models such as WaveNet~\cite{oordWaveNetGenerativeModel2016}. More recently, transformers have also been used for forecasting~\cite{zhouInformerEfficientTransformer2021}.
However, results in the literature including the Monash repository~\cite{godahewaMonashTimeSeries2021}, suggest that DNNs can still lag behind classic methods in simple forecasting benchmarks. We hypothesize that this gap can be reduced by improving the DNNs' input data distribution stability.

To boost performance on time series data, multiple attempts combined classic methods and neural networks, trying to overcome the weaknesses of both approaches. In \cite{zhangTimeSeriesForecasting2003}, authors propose a hybrid model made of an ARIMA and a neural network component. The M4 competition winner, the ES-RNN~\cite{smylHybridMethodExponential2020}, combines exponential smoothing and recurrent networks.
Similarly, GAS-Norm can be seen as a hybrid model that combines the GAS approach with a neural forecasting one.

Among the proposed statistical models for non-stationary data, Generalized Autoregressive Score models~\cite{crealGeneralizedAutoregressiveScore2013} are a class used for filtering time-varying parameters that update the estimation with the observed score. Their Information-Theoretic optimality has been shown with respect to the Kullback-Leibler divergence between the true and the estimated density function \cite{blasques2015information}. One example of a GAS model is the GARCH model, widely used to filter time-varying variance in time series data \cite{bollerslev1986generalized}.

Recently, different studies have tried to address challenges caused by non-stationary time series combining neural forecasting models and dynamic normalization procedures. 
Initial works focused only on the input normalization~\cite{ogasawara2010adaptive}. 
\cite{liAdaptiveBatchNormalization2018} proposes a batch normalization method for domain adaptation. DAIN~\cite{passalisDeepAdaptiveInput2019} learns the normalization with a nonlinear network, while \cite{deng2021stnorm} normalizes the input both in time and frequency. Unlike our proposal, these methods ignore non-stationarity over time within the input time series. RevIN~\cite{kimReversibleInstanceNormalization2021} introduced a denormalization step to restore the statistics removed during the normalization step. Similarly, \cite{fan2023dish} adopted a normalization methodology combined with a denormalization step. Despite this work considering also intra-space shift, i.e., non-stationarity between input and output time series, it adopts fixed statistics for the forecast.
Finally, SAN~\cite{liu2024san} proposed a dynamic normalization approach that splits both input and output into shorter temporal slices, in which non-stationarity can be less impactful, and uses them to estimate means and variances. Unlike SAN, our method adapts the statistics online and avoids possible problems caused by slices that are too long, like non-stationarity, or too short, such as overly noisy estimations.

\section{Proposed Method}
\subsection{Why Adaptive Normalization for a DNN?}
DNNs are a composition of many nonlinear functions, where the output of a layer becomes the input of the next one. Given the complex nonlinear structure of these models, the importance of the stability of the input distribution of each layer has been extensively explored in the literature \cite{klambauer2017self, ioffe2015batch}, showing improvements in training speed, stability, and generalization of DNNs. 

In this paper, we focus on improving the generalization of the nonlinear forecasting model in non-stationary settings. While DNNs can approximate any function \cite{lu2017expressive} (with some regularity conditions), their general definition also creates the risk of overfitting the observed data without learning the real generating function, particularly if the training data is noisy or 
does not fully represent the possible input space. Addressing this shortcoming is what our focus on generalization refers to, in the context of this paper. 

Using a simple example, we demonstrate that a complex nonlinear model can be less robust to changes in the input distribution than a linear model when applied to time series data. We generate a 3D Lorenz attractor as the solution of a Lorenz system with specified initial conditions, resulting in three chaotic time series representing the Cartesian coordinates of the system's evolution over time. Subsequently, we introduce a minimal amount of noise to these time series (see Figure \ref{subfig:Lorenz}).

We split the data into training, validation, and test sets, and train a simple 3-layer ReLU MLP to predict the future evolution of one coordinate based on the recent past of all three coordinates. This MLP is then compared to the same MLP using only linear activations, which is a linear model. Both models are tested on the same data shifted and rescaled with an affine transformation. Figure \ref{subfig:Ratio} shows the ratio between the test MSE of the nonlinear MLP and the MSE of the linear model. The plot reveals a "generalization area" (green area) where the nonlinear model outperforms the linear one. In this limited region, the nonlinear model remains superior even when the input has been shifted and rescaled. Outside of this region, the nonlinear model's performance rapidly degrades compared to the linear model (red area). For instance, when the data is shifted downward, the linear model significantly outperforms the ReLU MLP in predicting the chaotic time series.

In time series, changes in the mean and variance of the process are common and can be either predictable (like a simple linear trend) or unpredictable (like random regime changes)\cite{hamilton1994ftime}. In both cases, many approaches have been proposed, given the frequency of these kinds of problems. 
While our previous example shows the difficulties of a nonlinear model in case of an unexpected change in the location and scale of the data, limitations in learning can also arise due to predictable and deterministic trends. To illustrate this, we try to solve the same problem as in the previous example, but this time we add a quadratic trend to our time series. 

In this case, the nonlinear model is not able to learn the deterministic trend, resulting in an error higher than that of the linear model when the input data exits from the known distribution (Figure \ref{subfig:example_trend}). 

To solve these problems we need an adaptive and flexible way to filter the location and scale parameters of our data online. This would allow us to normalize the data even in the presence of deterministic or unpredictable non-stationarity.

\subsection{Parameter Filtering for Non-stationary Time Series}\label{sec:gas_theory}
\begin{figure}[]
    \centering
    \includegraphics[scale=0.098]{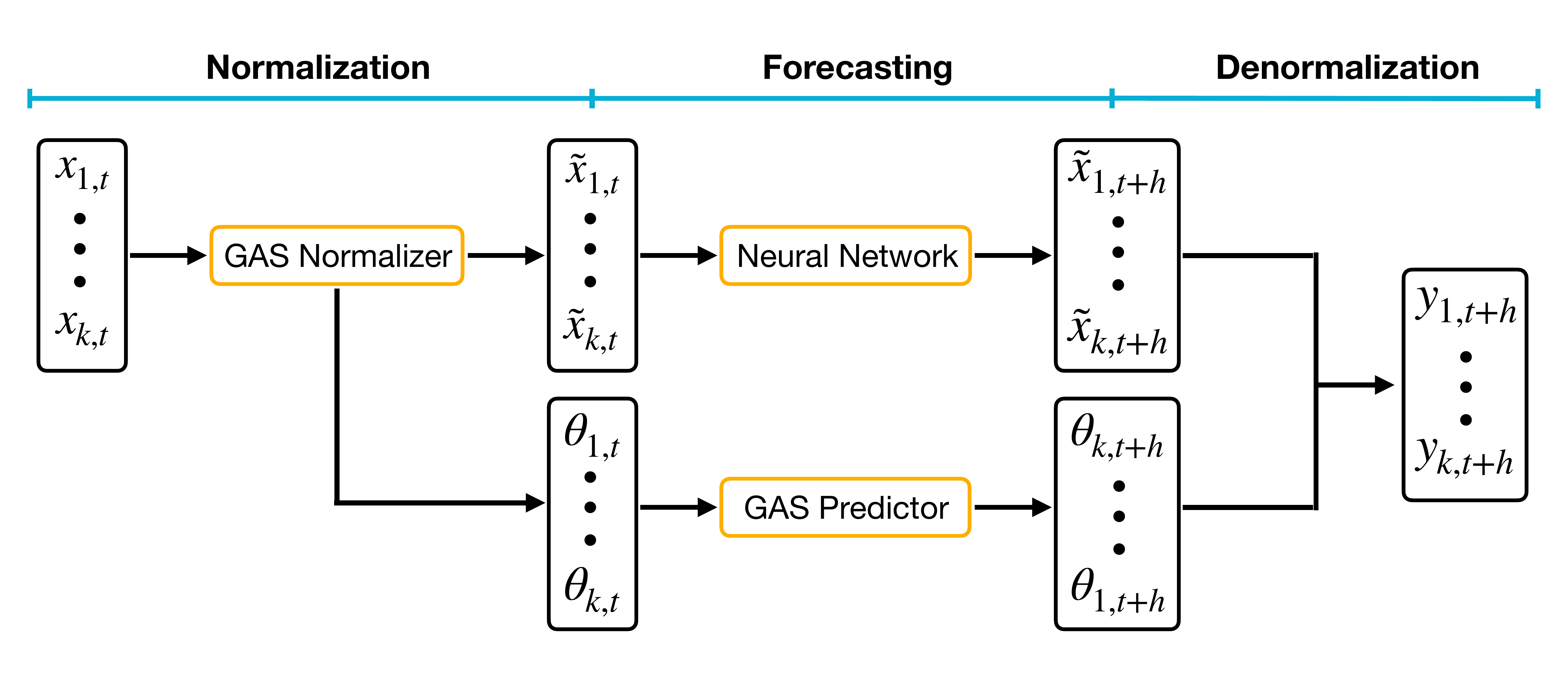}
    \caption{GAS-Norm procedure for a single time step input. Statistical parameters $\theta$ in this case are the means and variances.}
    \label{fig:enter-label}
\end{figure}
In this section, we define the theoretical framework of our filtering method.  
We set the forecasting problem as a conditional expectation problem where we predict, at time $t$, the expectation of a random variable $Y_{t+h}$ given the input $\{x_t, ..., x_{t-l}\}$, the ordered realizations of the random variables $\{X_t, ..., X_{t-l}\}$. All random variables can be multivariate. In common use, $Y_{t+h}$ and $\{X_t, ..., X_{t-l}\}$ are instances of the same stochastic process, where the first represents the future of the series and the second the past. It is also possible to use realizations from a process different from what we are predicting as input. The target of the forecasting problem is then:
$$E[Y_{t+h}|X_t, ..., X_{t-l}; w] = f_w(X_t, ..., X_{t-l})$$
The conditional expectation itself is a random variable that depends on the realizations of $\{X_t, ..., X_{t-l}\}$. The function $f_w$ is an arbitrary function parametrized by $w$. In deep learning, we try to approximate this function with a neural network $\hat{f}_w$, a strongly nonlinear model that is generally able to approximate any function \cite{hornik1989multilayer}. 

The input distribution at time $t$ of a $k$-dimensional input vector is an unknown joint distribution of all input features, denoted as $P_{X_t}(x_t)=P_{X_{1,t}, X_{2,t}, ... X_{k,t}}(x_{1,t}, x_{2,t}, ... x_{k,t})$, with mean vector $\mu_t = [\mu_{1,t}, \mu_{2,t}, ...\mu_{k,t}]$ and covariance matrix:
$$\Sigma_t = \begin{bmatrix} \sigma_{1,t}^2 & Cov(x_{1,t},x_{2,t}) & ...\\ Cov(x_{1,t},x_{2,t}) & \sigma_{2,t}^2 & ... \\ ... & ... & \sigma^2_{k,t}  \end{bmatrix}$$
The first index is the feature index. In the stationary case, the mean and the variance of the input distribution would be constant. In our non-stationary setting, they are both allowed to change in time. 

While keeping the input joint distribution unknown, we assume to know the type of parametrized density function of the marginal distributions of each input feature, conditional to the past observations of that feature itself. As an example, we can assume that the marginal distribution of each input feature is a Gaussian distribution with time-varying mean and variance:
$$
P_{X_{i,t} | \mathcal{F}_{i,t}}(x_{i,t}) = \mathcal{N}(\mu_{i,t}, \sigma_{i,t}^2)
$$
where $\mathcal{F}_{i,t}$ is the collection of observations of the feature $x_{i}$ that we have at time $t$.

To normalize each feature we need to filter the time-varying mean and variance. Given our parametric choice, we set our problem with an observation-driven state-space representation where the realizations of our input feature $x_{i,t}$ are given by:
\begin{align*}
    x_{i,t} &= \mu_{i,t} + \sigma_{i,t} \epsilon_{i,t}\\
    \mu_{i,t} &= g(\mu_{i,t-1}, x_{i,t-1}) \\
    \sigma^2_{i,t} &= g'(\sigma^2_{i,t-1}, x_{i,t-1}),
\end{align*} 
meaning that our input feature at time $t$ is generated by a random process with time-varying mean and variance and a random noise $\epsilon_{i,t}$ sampled from our assumed distribution. The value of the current states depends on their previous value and on the previous observation of the series.
To ease the notation, we use $\theta_t = [\mu_t, \sigma^2_t]$ as our time-varying parameter vector and we drop the feature index $i$. To solve the filtering problem, we need to find the best update for the parameter that maximizes the likelihood of the realizations, while penalizing changes that are too quick with respect to past predictions (as done in the Kalman Filter~\cite{bishop2006pattern}). In other words, we need a balance between the stability of the parameters and the update speed. As we will show, this is the problem GAS models try to solve. We modify the GAS formulation in \cite{lange2022robust} by adding a new hyperparameter $\gamma \in [0, 1)$ to control how much importance is given to maximizing the likelihood and how much is given to keeping the parameter stable. This hyperparameter explicitly controls the update speed of the normalization parameters, controlling the normalization strength of our method. Low values of $\gamma$ will result in a slower adaptation, leaving the normalized input more similar to the original one. At the extreme, $\gamma = 0$ is equivalent to a static normalization. The problem is written as:

$$\max_\theta \; \gamma \; \log p(x_{t} | \theta) - \frac{1-\gamma}{2} ||\theta - \theta_{t|t-1}||^2_{P_t}$$
where $P_t$ is a penalization matrix and $\theta_{t|t-1}$ is the prediction of the current parameter done at the previous time step. To compute explicitly the gradient, we approximate the problem with first-order Taylor expansion:
$$ \log p(x_{t} | \theta) =  \log p(x_{t} | \theta_{t|t-1}) + (\theta - \theta_{t|t-1}) \nabla_{\theta}(x_t | \theta_{t|t-1}).$$
The first-order condition to solve the maximization problem is:
$$\theta_{t|t} = \theta_{t|t-1} + \frac{\gamma}{1-\gamma} P_t^{-1} \; \nabla_{\theta}(x_t | \theta_{t|t-1}).$$

The penalization matrix proposed by \cite{crealGeneralizedAutoregressiveScore2013} is the Fisher Information Matrix (FIM), rescaled by a vector $\alpha^{-1}$ that works as a learning rate. The FIM is defined as the variance of the score and represents the expected curvature of the log-likelihood with respect to each parameter. By regularizing the score with the FIM, we obtain an online learning algorithm, where the parameter is updated at each new observation following the direction of the natural gradient $\tilde \nabla$. The natural gradient \cite{amari2016information} is the score adapted to the expected curvature of the log-likelihood. The final update step at each new observation is:
\begin{equation}
    \theta_{t|t} = \theta_{t|t-1} + \frac{\gamma}{1-\gamma} \alpha \; \tilde\nabla_{\theta}(x_t | \theta_{t|t-1})
\end{equation}
Finally, we add a simple linear prediction step guided by two parameters: $\omega$, which represents the unconditional mean of the parameters, and $\beta$, which controls the mean-reversion:
$$\theta_{t+1|t} = \omega + \beta \; \theta_{t|t}.$$
The whole statistics update process extends that of \cite{lange2022robust} by including an additional hyperparameter $\gamma$.
$\gamma$ can be selected with hyperparameters tuning techniques while the other parameters ($\alpha$, $\beta$, $\omega$) are directly optimized on the training data. To optimize these static parameters, we proceed with the prediction error decomposition where the time series of the single feature can be represented as:
\begin{equation*}\label{eq:this}
    p(x_1, x_2,...,x_T) = p(x_T | x_{T-1})... p(x_2 | x_1) p(x_1).
\end{equation*}
Maximizing the logarithm of Eq. \ref{eq:this} and adding a penalization term, we obtain a new optimization problem as a function of the static parameters:
\begin{align}
\begin{split}
    max_{\alpha, \omega, \beta} \; \; &\gamma \; \log p(x_1) - \frac{1-\gamma}{2} ||\theta_1 - \theta_0||^2_{P_t} + \\ 
    &\sum_{t=2}^T \gamma \; \log p(x_t|x_{t-1}) - \frac{1-\gamma}{2} ||\theta_t - \theta_{t|t-1}||^2_{P_t}.
\end{split}
\label{eq:param_optim}
\end{align}

Once we solve the optimization problem for each feature and we obtain the optimal $\alpha, \beta, \gamma$, we can use them to filter the mean and variance that will be used to normalize the marginal distribution of each feature at each time step. If the parameters are correctly filtered, after the normalization each marginal distribution has zero mean and unitary variance, and the joint input distribution becomes: 
\begin{align*}
    &P_{\tilde X_{1,t}, \tilde X_{2,t}, ... \tilde X_{k,t}}(\tilde x_{1,t}, \tilde x_{2,t}, ... \tilde x_{k,t})
    = \\ p([0, 0,...0], & \begin{bmatrix} 1 & Corr(x_{1,t},x_{2,t}) & ...\\ Corr(x_{1,t},x_{2,t}) & 1 & ... \\ ... & ... & 1  \end{bmatrix})
\end{align*}
an unknown distribution $p$ with mean zero and a covariance matrix equal to the correlation matrix of the original joint distribution. The \textit{tilde} symbol is used to denote the normalized feature. The normalized features can be used as input for the DNN model. 

\subsection{GAS-Norm}
\begin{figure}[t]
    \centering
    \includegraphics[scale=0.4]{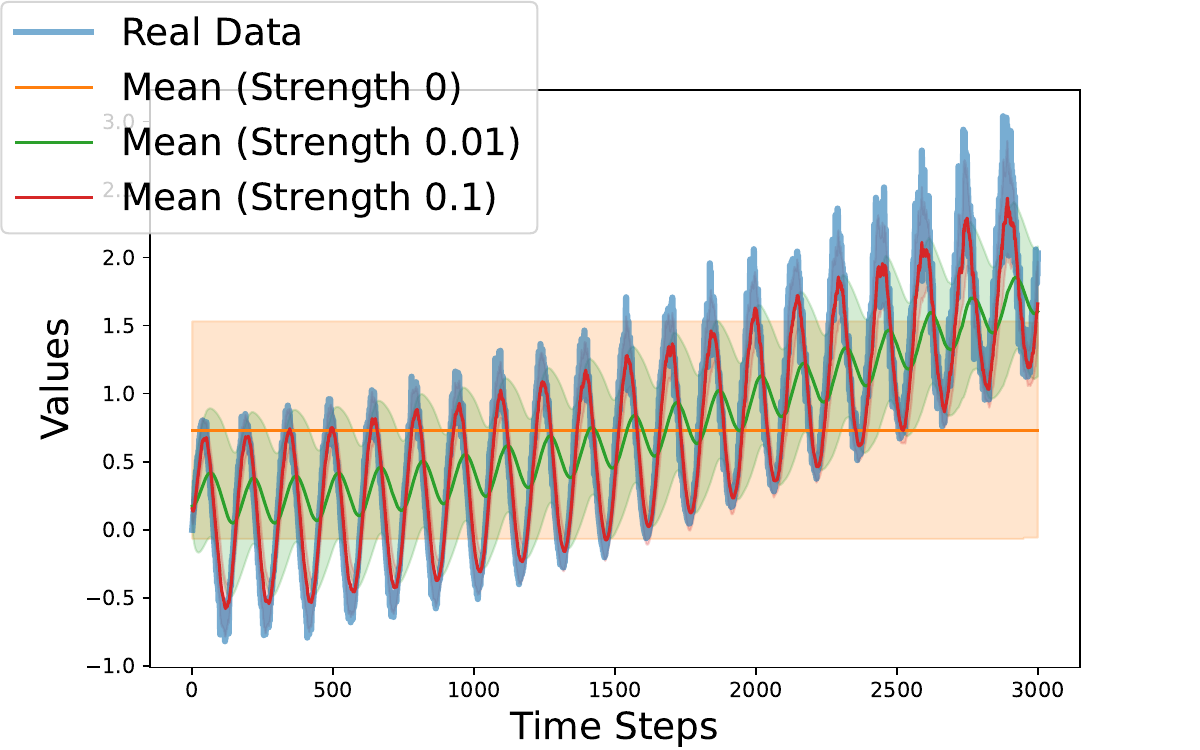}
    \caption{Comparison of the filtering process of GAS-Norm with different normalization strengths ($\gamma$). The shaded area is the mean $\pm$ standard deviation.}
    \label{fig:norm_strength}
\end{figure}
The GAS-Norm procedure proposed in this work is summarized in Figure \ref{fig:enter-label}, where different blocks show the normalization, forecasting, and denormalization phases, with input and output data for each of them. Formally, our forecasting model of the feature $y_{i, t+h}$ at a future time $t+h$ can be described as:
\begin{align}
\begin{split}
    y_{i,t+h} &= \mu_{i,t+h} + \sigma_{i,t+h} e_{i,t+h}\\
    \mu_{i,t+h} &= g(\mu_{i,t+h-1}) \\
    \sigma^2_{i,t+h} &= g'(\sigma^2_{i,t+h-1})\\
    e_{i,t+h} &= f_w(\tilde X_t, ...,  \tilde X_{t-l}) + \epsilon_{i, t+h},
\end{split}
\label{eq:de-norm}
\end{align}
where $e_{i,t+h}$ is now a deterministic residual and $\epsilon_{i, t+h}$ is random noise of unknown distribution\footnote{The distributional assumptions depend on the data characteristics.}.
The target of our forecasting problem, i.e., the expected value $E[Y_{t+h}|X_t, ..., X_{t-l}; w]$, is now the result of combining the independent prediction of the mean and variance process by GAS model, and the prediction done by the deep model using the normalized residuals. In this way, the DNN has the role of learning what the simple mean and variance predictions cannot forecast, effectively resulting in a \textbf{residual learning} procedure. 
Details of each step of our procedure are described in the following.
\begin{figure*}[t]
    \centering
    \parbox[t]{0.02\textwidth}{\centering}
    \parbox[t]{0.96\textwidth}{
        \begin{subfigure}[t]{0.3\textwidth}
            \centering
            \includegraphics[width=\textwidth]{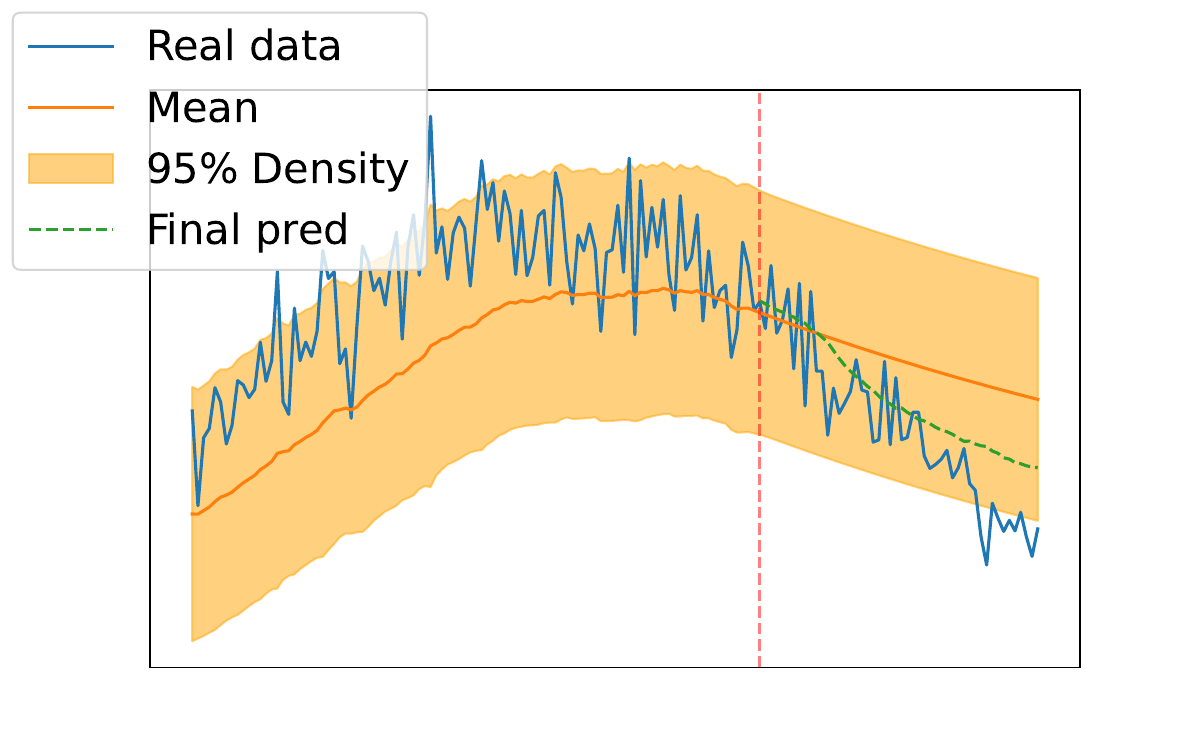}
        \end{subfigure}
        \hfill
        \begin{subfigure}[t]{0.3\textwidth}
            \centering
            \includegraphics[width=\textwidth]{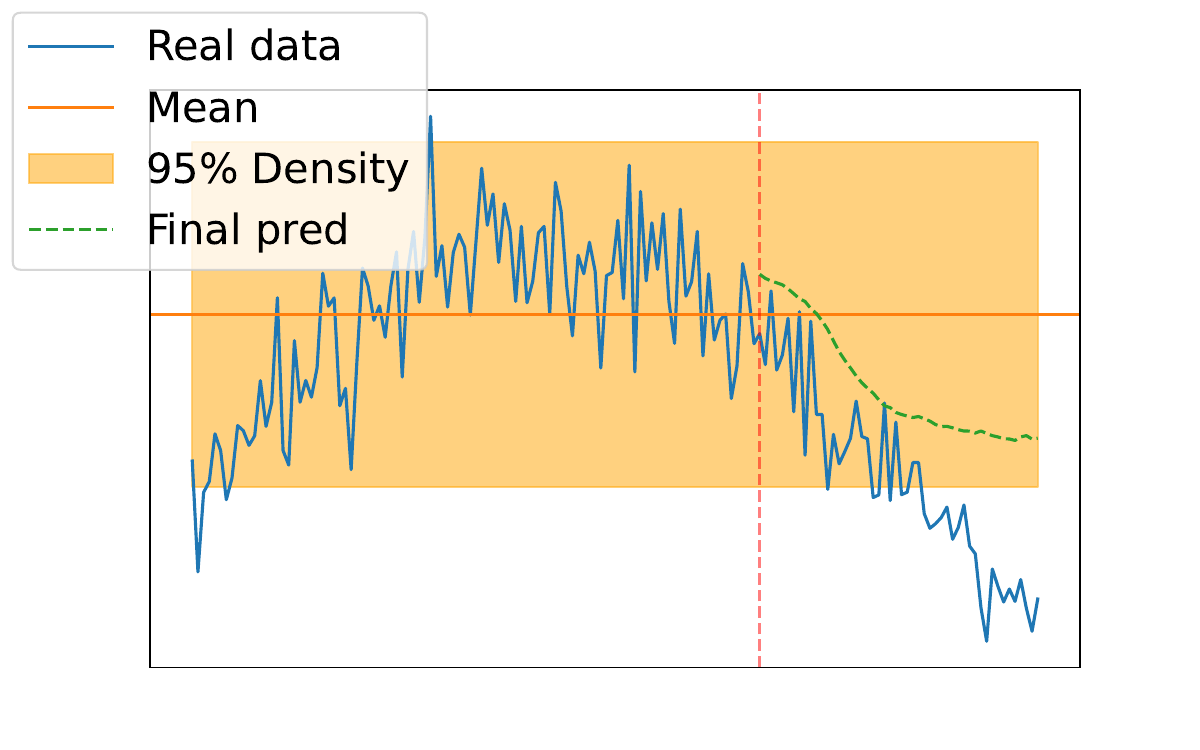}
        \end{subfigure}
        \hfill
        \begin{subfigure}[t]{0.3\textwidth}
            \centering
            \includegraphics[width=\textwidth]{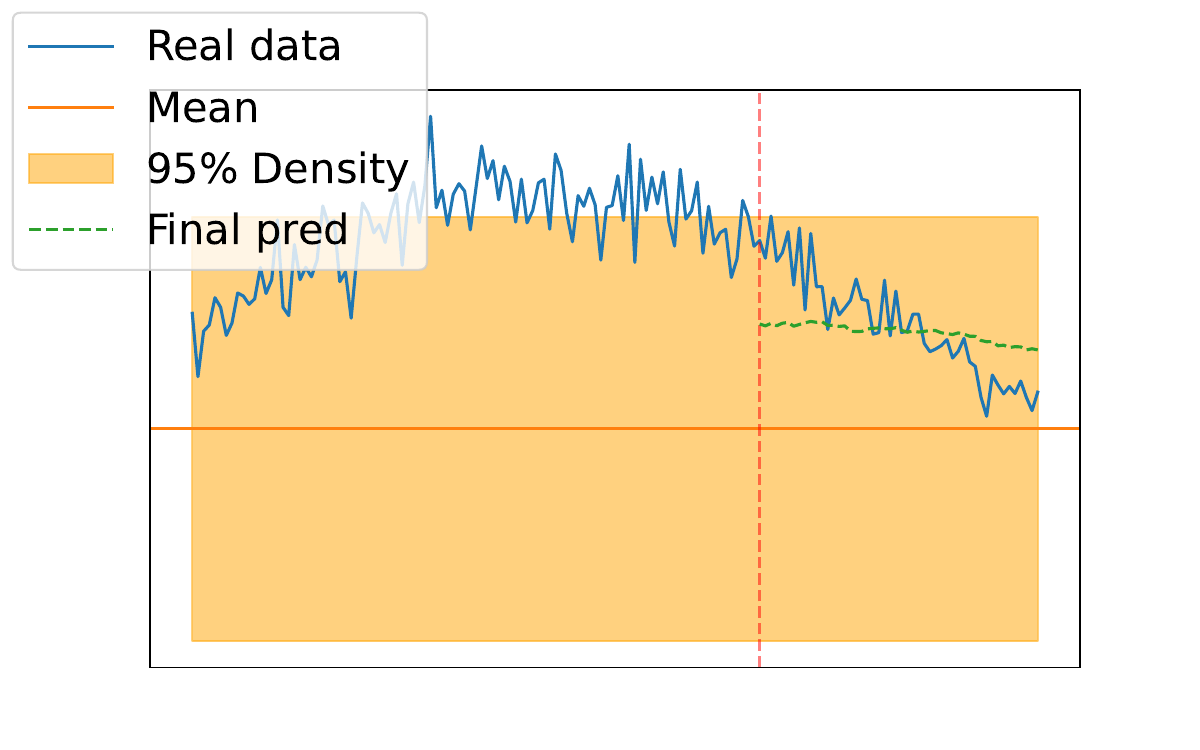}
        \end{subfigure}
    }

    \parbox[t]{0.02\textwidth}{\centering}
    \parbox[t]{0.96\textwidth}{
        \begin{subfigure}[t]{0.3\textwidth}
            \centering
            \includegraphics[width=\textwidth]{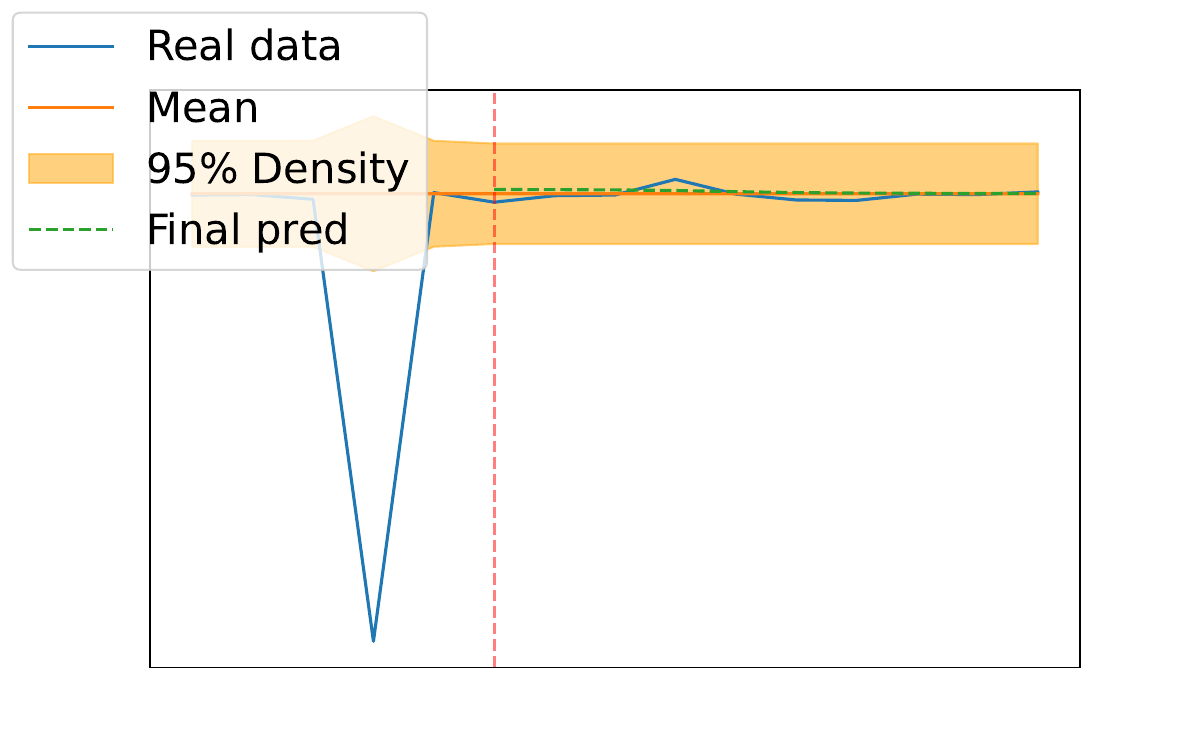}
            \caption{GAS-Norm}
        \end{subfigure}
        \hfill
        \begin{subfigure}[t]{0.3\textwidth}
            \centering
            \includegraphics[width=\textwidth]{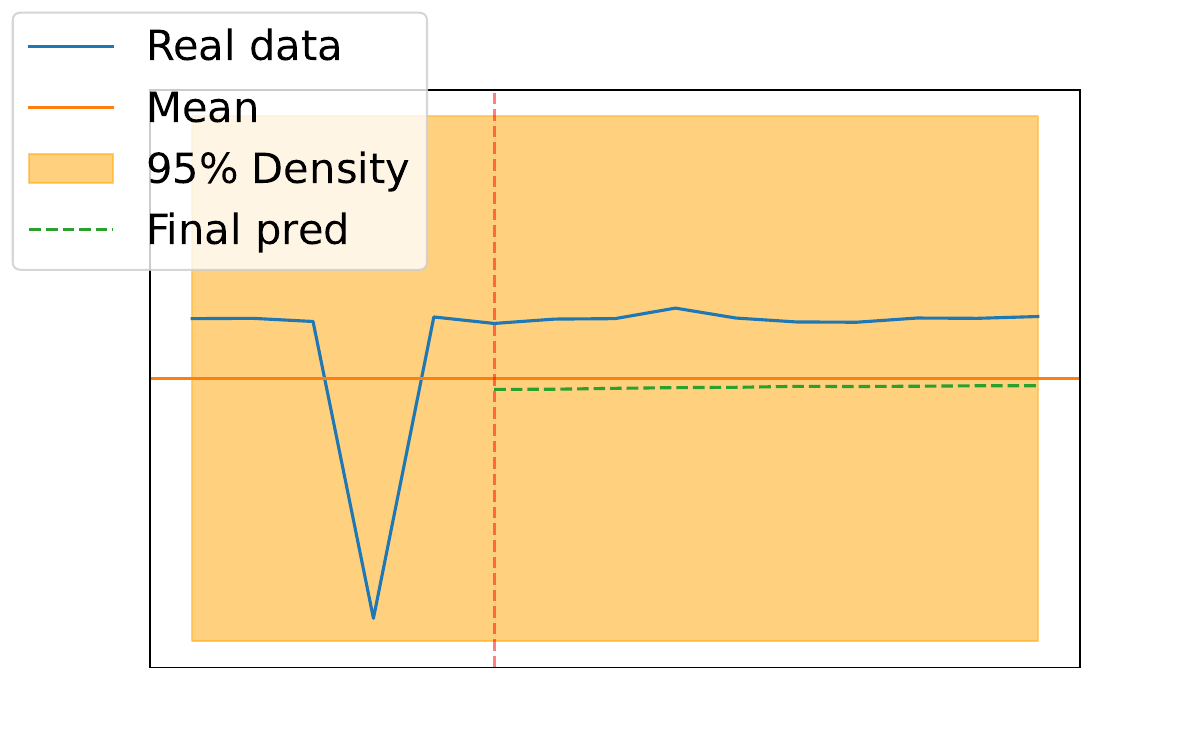}
            \caption{Local Norm}
        \end{subfigure}
        \hfill
        \begin{subfigure}[t]{0.3\textwidth}
            \centering
            \includegraphics[width=\textwidth]{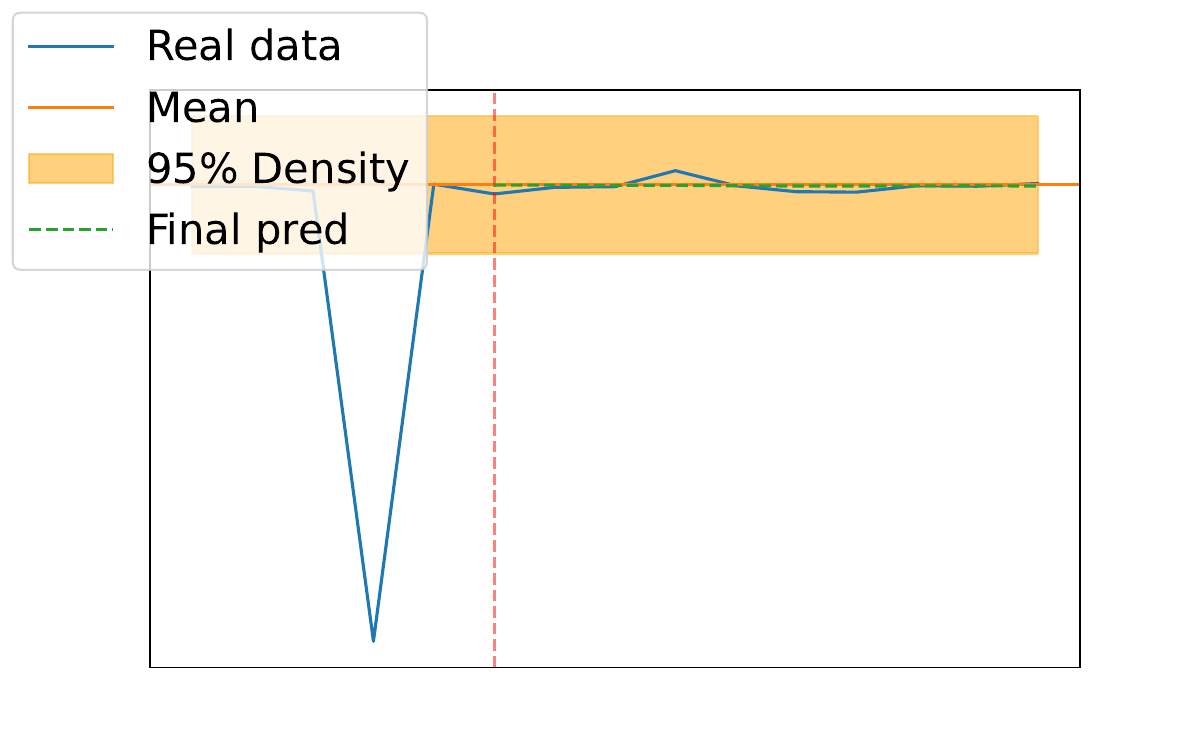}
            \caption{Global Norm}
        \end{subfigure}
    }

    \caption{Examples of input/output sequences with different methods' filtered and predicted statistics together with the denormalized prediction of the DNN. The top row shows a sequence from AR data, while the bottom row shows from the VIX data with an outlier. The red vertical dashed line splits the context part of the sequence from the part to be predicted.}
    \label{fig:enc/dec}
\end{figure*}

\smallskip
\noindent \textbf{Normalization.}
The normalization step is performed online on the input data (the portion of the input sequence used by the encoder).
The GAS model can be adapted to different distributional assumptions. We use a Student's t-distribution for our experiments. According to this choice, equations for mean and variance estimation (see the Appendix for the derivation) are computed as:
\begin{align}
\begin{split}
\mu_{t+1} =& \omega_\mu +  \beta_\mu [\frac{\gamma}{1-\gamma} \alpha_{\mu}\frac{y_t-\mu_t}{1 + \frac{(y_t - \mu_t)^2}{\nu \sigma^2_t}} + \mu_t]\\
    \sigma^2_{t+1}=& \omega_\sigma  + \beta_\mu [\frac{\gamma}{1-\gamma}\alpha_{\sigma}(\frac{(\nu + 1)(y_t - \mu_t)^2}{\nu + \frac{(y_t - \mu_t)^2}{\sigma^2_t}}-\sigma^2_t) +  \sigma^2_t]
\end{split}
\label{eq:studentt}
\end{align}
where, again, we removed the subscript $i$ from each variable to ease the notation.
As introduced in section \ref{sec:gas_theory}, the $\alpha, \beta$ and $\omega$ parameters result from the feature-wise optimization problem described in equation \ref{eq:param_optim}. The best values of these parameters are found for each feature of each time series of the training dataset, then they are left frozen during the training phase of the DNN.
Parameter $\nu$, which reflects the fatness of the tails of the Student's t-distribution, can be either optimized as the others or fixed. For large values of $\nu$, the Student's t converges to a Gaussian distribution. Finally, $\gamma$ is a hyperparameter selected with a validation set that controls the stability of the means and variances. It controls the amount of information about the original mean and variance the filtering process removes from the observed input sequence. In the rest of the paper, we will refer to this parameter also as normalization strength. 
The effect of different values of $\gamma$ can be seen in Figure \ref{fig:norm_strength}. Notice how, when setting the value to 0, our method collapses into a simple static normalization procedure, where the initial mean and variance can be set as the unconditional mean and variance of the whole training set.

This procedure for the online update of means and variances strives towards making the method more model-agnostic, while still providing stationary input to the deep model.

\smallskip
\noindent \textbf{Forecasting and Denormalization.} The normalized input time series is used by the deep model to make the forecast, leveraging the unfiltered linear and nonlinear correlations. At the same time, the GAS model is used to forecast future means and variances. In this case, the time series predicted by the models need to be part of the input features (the GAS model can only predict the same series it has filtered). We obtain the per-feature forecasting equation by modifying equations \ref{eq:studentt} substituting the observation with the last prediction of the mean and the variance:
\begin{align*}
    &\mu_{t+1} = \omega_\mu +  \beta_\mu\ \mu_{t}\\
    &\sigma^2_{t+1}= \omega_\sigma  + \beta_\sigma \sigma^2_{t}.
\end{align*}
This is a kind of autoregressive approach: the prediction is assumed as real observation to predict the next step.
Finally, the forecast made by the deep model is combined with the means and variances predicted by the GAS, hence re-introducing the information removed during the normalization procedure as in equation \ref{eq:de-norm}. During the training phase of the deep model, the final output is then used to compute the loss and update the DNN's parameters.

\begin{table*}[t]
    \centering
    \caption{Normalization methods comparison. Mean (standard error) of MASE on the test set.}
    \begin{tabular}{|l|l|l|l|l|l|l|l|}
    \hline
    \textbf{Data}          & \textbf{Enc length} & \textbf{GAS-Norm(100)} & \textbf{GAS-Norm(20)} & \textbf{Local Norm} & \textbf{Global Norm} & \textbf{BatchNorm} & \textbf{RevIN} \\ \hline
    \multirow{3}{*}{AR}  & 25             &   2.49 (0.459) & \textbf{2.0597 (0.3818)} & 2.0818 (0.1493) & 2.7052 (0.3712) & 4.9385 (0.4268) & 2.6416 (0.3685) \\

                         & 100            & \textbf{2.0038 (0.4591)} & 2.3813 (0.3722) & 2.3615 (0.289) & 2.5001 (0.2555) & 3.8886 (0.1987) & 3.4488 (0.0754) \\
                         & 200            & \textbf{1.9345 (0.3078)} & 2.1518 (0.1018) & 2.1728 (0.2807) & 2.3558 (0.2278) & 8.2509 (0.5197) & 5.3887 (0.4934) \\

 \hline
    \multirow{3}{*}{VIX} & 25       & 0.6841
 (0.0002)     &\textbf{0.6839 (0.0001)} & 0.6957 (0.0002) & 0.6843 (0.0001) & 0.6843 (0.0002) & 0.6958 (0.0002) \\
                         & 100     &  0.6734 (0.0002)
     &  \textbf{0.6733 (0.0001)} & 0.6758 (0.0002) & 0.6736 (0.0001) & 0.6735 (0.0002) & 0.6761 (0.0005) \\
                         & 200     &  0.6696 (0.0002)
     &\textbf{0.6695 (0.0002)} & 0.6699 (0.0003) & 0.6697 (0.0002) & 0.6696 (0.0001) & 0.6698 (0.0003) \\
 \hline
    \multirow{3}{*}{ECL} & 25             & 2.2159 (0.0377)
& \textbf{2.093 (0.0319)} & 2.1269 (0.0168) & 2.4096 (0.2796) & 4.5395 (0.0512) & 2.1266 (0.0156) \\
                         & 100            & 2.0942 (0.018) & 1.9876 (0.0181) & \textbf{1.9434 (0.0126)} & 2.311 (0.1835) & 4.4926 (0.0444) & 1.9437 (0.01) \\
                         & 200            & 2.2395 (0.1238)  & \textbf{1.9616 (0.0096)} & 2.0133 (0.0172) & 2.278 (0.2978) & 4.4318 (0.0422) & 2.0133 (0.0162) \\
 \hline
    \end{tabular}
    \label{tab:norm_comparison}
\end{table*}

\section{Experiments with Normalization Methods}
\label{sec:exp1}
In this section, we will focus on the comparison between the GAS-Norm and other normalization methods with data of different characteristics. We will begin with a qualitative comparison, followed by a quantitative experiment.  

Unlike other methods such as RevIN \cite{kimReversibleInstanceNormalization2021}, the EncoderNormalizer of the Pytorch-Forecasting library\footnote{\url{https://pytorch-forecasting.readthedocs.io/en/stable/}}, or Batch normalization \cite{ioffe2015batch}, our method is not dependent on the length of the encoder sequence or the batch size. The EncoderNormalizer (we will call it "Local Norm" as it is used also on GluonTS with another name) normalizes each input sequence using the mean and the variance of the single input itself, similarly to Instance normalization \cite{ulyanov2016instance}, by taking the mean and the variance from the single context length (the observed part of the series used as input for the encoder). RevIN extends this concept by adding a learnable affine transformation to the data. Both methods reverse their normalization for the output sequence with an inverse transformation (denormalization). Batch normalization instead, when applied to the input data, extracts the statistics from the batch dimension of the input and applies an affine transformation to the normalized data. These methods all depend on the particular training settings, losing their impact when the input sequence is too short (or the batch size too small) or too long (or too large). GAS-Norm, being directly fitted on the time series, does not suffer any of these limitations: by updating the mean and the variance used for normalizing the data at each timestep, our approach is truly adaptive and independent of the network architecture or training procedure. 

A more traditional approach consists of normalizing the data with the mean and the variance of the entire training set (we will call this method Global Norm). Another shortcoming of all these approaches is that, by normalizing a sequence with a single mean and variance, it will remain non-stationary. In contrast, GAS-Norm enables a high level of flexibility, partially removing the non-stationary behavior of the encoder sequence and allowing the use of a dynamic mean and variance to denormalize the final prediction of the deep model.  

\begin{table*}
\centering
\caption{MASE obtained with multiple architectures and normalization strategies on different datasets.}\label{table:res}
\begin{tabular}{|l|l|l|l|l|l|l|
}
\hline
\textbf{Dataset} & \textbf{Model} & \textbf{Default}          & \textbf{Local Norm} & \textbf{Global Norm}     & \textbf{Mean~ Scaling} & \textbf{GAS-Norm}         \\
\hline
NN5-8 Weekly       & FFNN           & \textbf{0.868 (0.041)} & 0.886 (0.019)    & 0.899 (0.033)          & 0.867 (0.032)       & 0.881 (0.030)          \\
                 & Transformer    & 1.138 (0.201)          & 1.589 (0.526)    & 0.895 (0.049)          & 0.952 (0.089)       & \textbf{0.828 (0.032)} \\
                 & DeepAR         & 1.043 (0.238)          & -                   & 0.868 (0.033)          & 1.048 (0.373)                        & \textbf{0.822 (0.031)}                           \\
                 & MQCNN          & 1.042 (0.054)          & 0.931 (0.027)    & 0.921 (0.008)          & 0.896 (0.069)       & \textbf{0.890 (0.013)} \\
\hline
NN5-35 Weekly       & FFNN           & 6.109 (7.914) & 1.389 (0.031)    & 1.409 (0.013)          & 4.774 (0.976)       & \textbf{1.274 (0.014)}          \\
                 & Transformer    & 1.753 (0.456)          & 1.697 (0.384)    & 1.349 (0.052)          & 1.638 (0.518)       & \textbf{1.309 (0.089)} \\
                 & DeepAR         & 1.663 (0.400)          & -                   & 1.369 (0.020)          & 1.837 (0.594)                        & \textbf{1.206 (0.018)}                           \\
                 & MQCNN          & 5.192 (3.462)          & 1.332 (0.053)    & 1.434 (0.031)          & 11.141 (4.630)       & \textbf{1.291 (0.066)} \\
\hline
M4 Weekly        & FFNN           & 0.603  (0.012)          & 0.582  (0.008)    & 0.591  (0.017)          & 0.620  (0.007)       & \textbf{0.572  (0.009)} \\
                 & Transformer    & 3.674  (0.190)          & 1.689  (0.046)    & 0.952  (0.142)          & 0.872  (0.046)       & \textbf{0.685  (0.059)} \\
                 & DeepAR         & 1.528  (0.119)          & -                   & \textbf{0.603  (0.036)}          & 1.643  (0.105)                       & 0.606  (0.018)                          \\
                 & MQCNN          & 0.751  (0.051)          & 0.683  (0.033)    & 0.778  (0.041)          & 0.809  (0.037)       & \textbf{0.673  (0.026)} \\
\hline
Fred MD          & FFNN           & 0.765 (0.054)          & 0.657 (0.048)    & 0.593(0.024)          & 0.741 (0.034)       & \textbf{0.593 (0.030)} \\
                 & Transformer    & 8.411 (0.339)          & 2.360 (0.028)    & 0.733 (0.053)          & 1.459 (0.496)       & \textbf{0.719 (0.112)} \\
                 & DeepAR         & 5.540 (0.091)          & -                   & \textbf{0.629 (0.065)} & 6.480 (0.130)                       & 0.773 (0.193)          \\
                 & MQCNN          & 0.776 (0.076)          & 0.813 (0.010)    & 0.768 (0.080)          & 0.821 (0.046)       & \textbf{0.662 (0.050)} \\
\hline           
\end{tabular}
\end{table*}

Figure \ref{fig:enc/dec} illustrates some examples of how GAS-Norm, Local Norm, and Global Norm will act on an encoder/decoder sequence. The first row shows an example of AR data (described below). It is clearly visible how GAS-Norm is able to follow the data, while also helping the deep model by providing a prediction for the future behavior of the sequence. The Global Norm, by using a single global statistic for all the data, is visibly using a wrong variance to standardize this sequence. 

In the second row, the same comparison is done with an example of VIX data\footnote{\url{www.cboe.com/tradable_products/vix/}} (described below) with the addition of an outlier. In this case, the Global Norm's robustness is playing in its favor. A single outlier will not change the overall mean and variance computed on the entire training set. Conversely, the Local Norm will be highly affected by an outlier (as would RevIN), particularly in the case of shorter input sequences. GAS-Norm, instead, can assume any distribution, becoming robust to outliers. In this case, a Student's t-distribution with 20 degrees of freedom is assumed, thus mitigating the large changes encouraged by the outliers. 

\subsection{Quantitative Evaluation}

To compare the different normalization methods, we test them in three widely different univariate datasets. In our experiments, we compare GAS-Norm to standard normalization approaches like Global Norm and batch normalization \cite{ioffe2015batch} and to time series-specific approaches like Local Norm and the well-known state-of-the-art RevIN \cite{kimReversibleInstanceNormalization2021}. The experiments are performed in three datasets with different characteristics:
\begin{enumerate}
    \item AR: The AR dataset is a synthetic dataset generated by the built-in function of the pytorch-forecasting library "generate-ar-data". It is the realization of an autoregressive process to which a sinusoidal seasonality and a trend are added. This data is particularly difficult to learn due to the constantly moving mean of the input data. 

    \item VIX: The Chicago Board Options Exchange's Volatility Index is a fundamental financial index measuring the expected volatility of the stock market, implied in the S\&P 500 index options. We used the daily data from January 1990 to May 2024 downloaded from Yahoo Finance. As is frequently done in econometrics \cite{box2015time}, we took the first difference of the data thus removing the non-stationarity of the mean. The resulting series is highly heteroscedastic, with an increase in the variance during financial crises and a decrease in variance during stable times. 

    \item ECL: The Electricity Consuming Load\footnote{\url{archive.ics.uci.edu/dataset/321}} contains the hourly kWh electricity consumption of multiple clients. We took a subset of it (the first 10 thousand hours of the first client). This data has strong recurrent patterns due to habit repetition and sudden jumps between high and low consumption. 
\end{enumerate}

We apply the different normalization methods to a two-layer LSTM, always predicting the next 50 steps in time but using different input sequence lengths. GAS-Norm is used assuming a Student's t-distribution experimented with 20 and 100 degrees of freedom. Hyperparameter optimization is performed for the learning rate of each method and the normalization strength of GAS-Norm.
Each configuration is trained 10 times with different random seeds. The training is executed on a Tesla V100 16GB. We report the Mean Absolute Scaled Error (MASE) for all our experiments. 
The fundamental work of Hyndman, et al. \cite{hyndman2006another} suggests this as the preferable metric to compare forecasting models on different series. Note that a value less than 1 is acceptable in multistep forecasting due to the additional complexity of predicting a distant future compared to the easier naive approach with access to the last time step. The means and the standard deviations of the MASE in the test set are presented in Table \ref{tab:norm_comparison}. In our tests, GAS-Norm showed remarkable robustness to the different data characteristics. AR data, since it is synthetic, has no outliers, resulting in the GAS-Norm with 100 degrees of freedom winning the comparison with both the Local Norm and the Global Norm performing similarly well. Notice also how the longer the input sequence, the better the Global Norm performs. VIX data is very heteroscedastic with many outliers. As expected, the robust GAS-Norm performs better. With the ECL data, both RevIN and the Local Norm obtain good results compared to the other methods (this is one of the datasets used in RevIN paper \cite{kimReversibleInstanceNormalization2021}). Still, GAS-Norm is the best method in 2 out of 3 settings. Since ECL shows strong seasonality with sudden jumps, GAS-Norm results could be greatly improved by using a seasonal GAS, which we leave as future work. 

\section{Experiments with SOTA Forecasting Models}

In addition to the above comparisons, we also evaluate our method using state-of-the-art forecasting models on more extended real-world data.
For these experiments, we rely on the Monash Forecasting repository \cite{godahewaMonashTimeSeries2021} and their model evaluation procedure. Datasets in the repository are collections of several time series. Each time series shares the same context length and prediction length. The former refers to the length of the portion of the time series to be used as input (sometimes known as the lag length). The latter is the forecasting horizon. Datasets are then used to train a global probabilistic forecasting model, i.e., a single model trained across all series to predict the parameters of a pre-defined output distribution (Student's t in our case). To evaluate the models, a portion of elements from the end of each time series are used as the test set, and the length of that portion is equal to the prediction length.

We selected three different datasets, Fred MD, NN5 weekly, and M4 weekly, which present different non-stationary natures and cover a variety of domains. Table \ref{table:lags_and_horizons} shows some statistics for each dataset, including context and prediction lengths used. We use the same lengths for these datasets as the Monash repository curators. In the case of NN5 weekly, we also test our procedure with a prediction length equal to 35, the longest possible length for this dataset.
The GluonTS package for time series modeling \cite{alexandrov2020gluonts} is used as the foundation for the deep learning models. Among these are the feedforward neural network \cite{rumelhart1986learning}, transformer \cite{vaswani2017attention}, DeepAR \cite{salinas2020deepar}, and MQCNN \cite{wen2017multi} models. These models are chosen as they represent various complexities and are based on the most common deep learning architectures. 

We extended GluonTS to support different normalization methods. Training is again done using a Tesla V100 with 16GB of RAM.
Each model is trained and evaluated with GAS-Norm, global and local normalization settings from section \ref{sec:exp1} and two additional procedures:
\begin{itemize}
    \item Default state: The out-of-the-box state of the model's implementation. For DeepAR, this state includes local normalization. For the other models, it includes no normalization.
    \item Mean scaling: dividing values in the time series using the average value of the previous context length values.
\end{itemize}
Since the model forecasts are distributional, the output is taken to be the mean of 100 forecast samples.
For each approach, we perform the same hyperparameter tuning on each model.
The hyperparameters were tuned over 10 trials, selecting the best configuration on the validation set for the final evaluation. The tuning search spaces are shown in Table~\ref{table:tuning_space}. 

In addition to searching the model hyperparameters, we also perform a simple search for the GAS-Norm hyperparameter, i.e., the normalization strength. The search space is (0, 0.001, 0.01, 0.1, 0.5). For each strength value, we repeat the model hyperparameter tuning as described above, thus resulting in the selection of the normalization strength, along with the corresponding model hyperparameters.

After selecting the best values, they are used to train and evaluate each model five times, recording the mean absolute scaled error results.

\subsection{Results}
The mean and standard deviation of these results are presented in Table~\ref{table:res}.
In 13 out of 16 configurations, the GAS-Norm approach gives the best result.
Global normalization is the most effective in 2 occurrences, but it is not meaningfully ahead of GAS-Norm. In one unusual case of the default setups, for NN5-8 weekly with FFN, there is an occurrence where it had the best result so any added normalization worsens performance. However, considering the small differences across settings for that model and dataset, the occurrence could be due to the limited number of trials and tuning. Furthermore, when the default setups perform poorly they show results notably worse than the rest. Mean scaling, on the other hand, shows somewhat similar behavior as the default setups. That is, it suffered in the same cases as the defaults, but without excelling in any configurations. For local normalization, DeepAR was excluded from this evaluation since its default state already includes it. The remainder of configurations for this approach did not stand out in performance but contained no unusually poor instances either. An important caveat to note with these experimental results is that this GAS-Norm implementation is only a starting point for the method's capabilities. It is probably possible to get additional improvements by leveraging time series seasonality or using tailored distributional assumption. 

Comparing the different forecasting methods, the results show that DeepAR and FFN tend to outperform the rest on two datasets each. On the other hand, the worst results for each dataset came from the Transformer on three benchmarks and MQCNN on one. We hypothesize that internal normalization methods in the Transformer and MQCNN (LayerNorm or BatchNorm layers), which do not include a denormalization step, may be partially responsible for the underperformance.

The GAS-Norm's normalization phase added a negligible overhead. More specifically, with an Intel Xeon Gold 6140M CPU, the normalization phases for NN5 weekly, M4 weekly and Fred MD took 4.08s, 28.46s and 8.37s respectively at a mean and variance strength of 0. The strength parameter affects the runtime of the normalization phase, where they increase in tandem. So the same respective datasets with a strength of 0.5 yielded runtimes of 14.55s, 61.08s and 307.02s. 

\section{Conclusion}
In this paper, we show that non-stationarity is one of the main challenges when training nonlinear deep networks for forecasting. We focus on non-stationarity in the mean and variance, showing that they disrupt the predictions of nonlinear deep networks even in simple settings. To mitigate this issue, we propose GAS-Norm, a novel normalization method. The GAS model works as a filter that updates the mean and variance online. As a result, the normalized input passed to the deep network is more stable even in non-stationary settings. The output of the deep network is denormalized using the GAS predictions, obtaining the final forecasting. We evaluate GAS-Norm on a wide array of diverse datasets and models, which encompass the most popular modeling choices. We show that this method is more robust than popular normalization methods both on synthetic data and in real benchmarks. The results also show that most forecasting methods improve when combined with GAS-Norm.

GAS-Norm provides a general framework that can be easily extended in future works with different distributional assumptions and update dynamics, such as counting processes or time series with explicit seasonality. Additionally, it could also be investigated as a way to mitigate non-stationarity of the hidden activations of the network.

\appendix

\section{Appendix}
\subsection{Data Statistics and Hyperparameters}
\begin{table}[h!]
    \centering
    \caption{Hyperparameter tuning space per model.} \label{table:tuning_space}
    \begin{tabular}{|l|l|l|} 
    \hline
    \textbf{Model}                                                         & \textbf{Hyperparameter} & \begin{tabular}[c]{@{}l@{}}\textbf{Searched}\\\textbf{Values}\end{tabular}  \\ 
    \hline
    Feed Forward                                                           & Number of Layers        & {[}1,5]                                                                     \\ 
    \hline
                                                                           & Hidden Dimensions       & {[}10,100]                                                                  \\ 
    \hline
                                                                           & Learning Rate           & {[}1e-3,1e-6]                                                               \\ 
    \hline
                                                                           & Number of Epochs        & {[}10,100]                                                                  \\ 
    \hline
    \begin{tabular}[c]{@{}l@{}}Transformer,\\ DeepAR,\\ MQCNN\end{tabular} & Learning Rate           & {[}1e-3,1e-6]                                                               \\ 
    \hline
                                                                           & Number of Epochs        & {[}10,100]                                                                  \\
    \hline
    \end{tabular}
\end{table}

\begin{table}[h!]
    \centering
    \caption{Dataset statistics.} \label{table:lags_and_horizons}
    \begin{tabular}{|l|l|l|l|l|} 
    \hline
    \textbf{Dataset} & \begin{tabular}[c]{@{}l@{}}\textbf{\# Time}\\\textbf{Series}\end{tabular} & \begin{tabular}[c]{@{}l@{}}\textbf{Avg TS}\\\textbf{Length}\end{tabular} & \begin{tabular}[c]{@{}l@{}}\textbf{Context}\\\textbf{Length}\end{tabular} & \begin{tabular}[c]{@{}l@{}}\textbf{Prediction}\\\textbf{Length}\end{tabular}  \\ 
    \hline
    Fred MD          & 107 & 728 & 15                                                                        & 12                                                                            \\ 
    \hline
    NN5 Weekly       & 111 & 113 & 65                                                                        & 8, 35                                                                             \\ 
    \hline
    M4 Weekly        & 359 & 1035 & 26                                                                        & 13                                                                            \\
    \hline
    \end{tabular}
\end{table}

\subsection{Gaussian Score Driven Adaptive Normalization}

In this section, we show the Gaussian version of the GAS normalization procedure.

Assume each feature to be a time series with a Gaussian conditional distribution:
\begin{equation*}
    y_{i,t} \sim \mathcal{N}(\mu_{i,t}, \sigma^2_{i,t})
\end{equation*} 
Where $i$ is the feature index that we can now drop, considering each feature separately with the same procedure. 
The update for the two parameters of the distribution becomes:
\begin{align*}
    \begin{pmatrix}
        \mu_{t+1} \\ \sigma^2_{t+1}
    \end{pmatrix}
    = &
    \begin{pmatrix}
        \omega_\mu \\ \omega_\sigma 
    \end{pmatrix}
    +
    \begin{pmatrix}
        \frac{\gamma}{1-\gamma} \alpha_\mu & 0 \\ 0 &\frac{\gamma}{1-\gamma} \alpha_\sigma 
    \end{pmatrix}
    S_t
    \begin{pmatrix}
        \nabla_\mu \\ \nabla_\sigma
    \end{pmatrix}
    + \\
     &\begin{pmatrix}
        \beta_\mu & 0 \\ 0 &\beta_\sigma 
    \end{pmatrix}
    \begin{pmatrix}
        \mu_{t} \\ \sigma^2_{t}
    \end{pmatrix}
\end{align*} 
where $S_t$ is a $2\times2$ scaling matrix (we use the Fisher Information Matrix as suggested by Creal et al. \cite{crealGeneralizedAutoregressiveScore2013} and to obtain a natural gradient \cite{amari2016information} adapted to the data geometry). We now compute the scores for the mean, the variance and the component of the FIM: 
\begin{align*}
    &log \mathcal{N}(y_t | \mu_t, \sigma^2_t) = -\frac{1}{2}log(2 \pi) - \frac{1}{2}log \sigma_t^2 - \frac{1}{2 \sigma_t^2} (y_t - \mu_t)^2 \\
    &\frac{\partial log \mathcal{N}(y_t | \mu_t, \sigma^2_t)}{\partial \mu_t}  = \frac{y_t - \mu_t}{\sigma_t^2}\\
    &\frac{\partial log \mathcal{N}(y_t | \mu_t, \sigma^2_t)}{\partial \sigma^2_t} = \frac{1}{2}( \frac{(y_t - \mu_t)^2}{\sigma^4}-\frac{1}{\sigma_t^2})\\
    &\mathbf{V}[\frac{\partial log \mathcal{N}(y_t | \mu_t, \sigma^2_t)}{\partial \mu_t}] = \frac{1}{\sigma^2}\\
    &\mathbf{V}[\frac{\partial log \mathcal{N}(y_t | \mu_t, \sigma^2_t)}{\partial \sigma^2_t}] = \frac{1}{2\sigma^4} 
\end{align*}
The covariance between the two gradients is $0$. Given our parameterization of the update, this results in two independent update functions. 
Using the Fisher Information as a scaling matrix we obtain:
\begin{align*}
    &\mu_{t+1 | t} = \omega_\mu +  \beta_\mu[\frac{\gamma}{1-\gamma} \alpha_{\mu}(y_t-\mu_t) + \mu_t]\\
    &\sigma^2_{t+1 | t}= \omega_\sigma  + \beta_\sigma[\frac{\gamma}{1-\gamma} \alpha_{\sigma} ((y_t-\mu_t)^2-\sigma^2_t) + \sigma^2_t].
\end{align*}

\subsection{Student's t Score Driven Adaptive Normalization}
Student's t-distribution has a subexponential decay rate, allowing for fat tails. This means that this distribution can be used when we expect to have outliers in our feature time series. The degrees of freedom control the shape of the distribution, with fatter tails with lower degrees of freedom.

Again we standardize the features independently from the others. We assume a feature $i$ to be generated as:
\begin{align*}
    y_t = \mu_t + \sigma_t \epsilon_t && \epsilon_t \sim t(0, 1, \nu)
\end{align*}
where $\nu$ represents the degrees of freedom. 

The resulting likelihood is
\begin{equation*}
    t(y_t | \mu_t, \sigma^2_t, \nu) = \frac{\Gamma (\frac{\nu + 1}{2})}{\Gamma(\frac{\nu}{2}) \sqrt{\pi \nu \sigma^2}} (1 + \frac{(y_t - \mu_t)^2}{\nu \sigma^2})^{-(\nu+1)/2},
\end{equation*}
where $\Gamma$ is the Gamma function.
The corresponding log-likelihood becomes
\begin{align*}
     log t(y_t | \mu_t, \sigma^2_t, \nu) = & log (\frac{\Gamma (\frac{\nu + 1}{2})}{\Gamma(\frac{\nu}{2}) \sqrt{\pi \nu}}) - \frac{1}{2} log (\sigma^2) - \\ 
     & - \frac{\nu + 1}{2} log(1 + \frac{(y_t - \mu_t)^2}{\nu \sigma^2}).
\end{align*}

We can now compute the gradients that compose the score and the variances of the scores:
\begin{align*}
    &\frac{\partial t(y_t | \mu_t, \sigma^2_t, \nu)}{\partial \mu_t} = \frac{(\nu + 1)(y_t - \mu_t)}{\nu \sigma^2 + (y_t - \mu_t)^2} \\
  &\frac{\partial t(y_t | \mu_t, \sigma^2_t, \nu)}{\partial \sigma^2_t} = \frac{1}{2}(\frac{(\nu + 1)(y_t-\mu_t)^2}{\nu \sigma^4 + \sigma^2 (y_t-\mu_t)^2} -\frac{1}{\sigma^2}) \\
  &V[\frac{\partial t(y_t | \mu_t, \sigma^2_t, \nu)}{\partial \mu_t}] = \frac{\nu + 1}{(\nu+3)\sigma^2_t}\\
  &V[\frac{\partial t(y_t | \mu_t, \sigma^2_t, \nu)}{\partial \sigma^2_t}] = \frac{\nu}{2 (\nu+3) \sigma^4_t}.
\end{align*}

Again the correlation between the two gradients is $0$. The degrees of freedom are chosen as a static parameter. The dynamic version is possible but the optimization is much more difficult. As suggested by Artemova et al. \cite{artemova2022score}, we regularize scores with a scaling proportional to the inverse Fisher information: $S_{\mu,t}=\frac{\nu \sigma^2_t}{1+\nu}$ for the mean and $S_{\mu,t}=2\sigma^4$ for the variance:

The update we obtain is:
\begin{align*}
    \mu_{t+1} =& \omega_\mu +  \beta_\mu [\frac{\gamma}{1-\gamma} \alpha_{\mu}\frac{y_t-\mu_t}{1 + \frac{(y_t - \mu_t)^2}{\nu \sigma^2_t}} + \mu_t]\\
    \sigma^2_{t+1}=& \omega_\sigma  + \beta_\mu [\frac{\gamma}{1-\gamma}\alpha_{\sigma}(\frac{(\nu + 1)(y_t - \mu_t)^2}{\nu + \frac{(y_t - \mu_t)^2}{\sigma^2_t}}-\sigma^2_t) +  \sigma^2_t]
\end{align*}

\begin{acks}
Antonio Carta acknowledges support from the Ministry of University and
Research (MUR) as part of the FSE REACT-EU - PON 2014-2020 “Research and
Innovation" DM MUR 1062/2021. Reshawn J. Ramjattan acknowledges support from the Ministry of University and Research (MUR) as part of the PON 2014--2020 ``Research and Innovation'' resources---Green/Innovation Action---DM MUR 1061/2022.
\end{acks}

\bibliographystyle{ACM-Reference-Format}
\bibliography{main}


\begin{thebibliography}{43}


\ifx \showCODEN    \undefined \def \showCODEN     #1{\unskip}     \fi
\ifx \showDOI      \undefined \def \showDOI       #1{#1}\fi
\ifx \showISBNx    \undefined \def \showISBNx     #1{\unskip}     \fi
\ifx \showISBNxiii \undefined \def \showISBNxiii  #1{\unskip}     \fi
\ifx \showISSN     \undefined \def \showISSN      #1{\unskip}     \fi
\ifx \showLCCN     \undefined \def \showLCCN      #1{\unskip}     \fi
\ifx \shownote     \undefined \def \shownote      #1{#1}          \fi
\ifx \showarticletitle \undefined \def \showarticletitle #1{#1}   \fi
\ifx \showURL      \undefined \def \showURL       {\relax}        \fi
\providecommand\bibfield[2]{#2}
\providecommand\bibinfo[2]{#2}
\providecommand\natexlab[1]{#1}
\providecommand\showeprint[2][]{arXiv:#2}

\bibitem[Alexandrov et~al\mbox{.}(2020)]%
        {alexandrov2020gluonts}
\bibfield{author}{\bibinfo{person}{Alexander Alexandrov}, \bibinfo{person}{Konstantinos Benidis}, \bibinfo{person}{Michael Bohlke-Schneider}, \bibinfo{person}{Valentin Flunkert}, \bibinfo{person}{Jan Gasthaus}, \bibinfo{person}{Tim Januschowski}, \bibinfo{person}{Danielle~C Maddix}, \bibinfo{person}{Syama Rangapuram}, \bibinfo{person}{David Salinas}, \bibinfo{person}{Jasper Schulz}, {et~al\mbox{.}}} \bibinfo{year}{2020}\natexlab{}.
\newblock \showarticletitle{Gluonts: Probabilistic and neural time series modeling in python}.
\newblock \bibinfo{journal}{\emph{Journal of Machine Learning Research}} \bibinfo{volume}{21}, \bibinfo{number}{116} (\bibinfo{year}{2020}), \bibinfo{pages}{1--6}.
\newblock


\bibitem[Amari(2016)]%
        {amari2016information}
\bibfield{author}{\bibinfo{person}{Shun-ichi Amari}.} \bibinfo{year}{2016}\natexlab{}.
\newblock \bibinfo{booktitle}{\emph{Information geometry and its applications}}. Vol.~\bibinfo{volume}{194}.
\newblock \bibinfo{publisher}{Springer}.
\newblock


\bibitem[Artemova et~al\mbox{.}(2022)]%
        {artemova2022score}
\bibfield{author}{\bibinfo{person}{Mariia Artemova}, \bibinfo{person}{Francisco Blasques}, \bibinfo{person}{Janneke van Brummelen}, {and} \bibinfo{person}{Siem~Jan Koopman}.} \bibinfo{year}{2022}\natexlab{}.
\newblock \showarticletitle{Score-driven models: Methodology and theory}.
\newblock In \bibinfo{booktitle}{\emph{Oxford Research Encyclopedia of Economics and Finance}}.
\newblock


\bibitem[Bishop(2006)]%
        {bishop2006pattern}
\bibfield{author}{\bibinfo{person}{Christopher~M Bishop}.} \bibinfo{year}{2006}\natexlab{}.
\newblock \showarticletitle{Pattern recognition and machine learning}.
\newblock \bibinfo{journal}{\emph{Springer google schola}}  \bibinfo{volume}{2} (\bibinfo{year}{2006}), \bibinfo{pages}{1122--1128}.
\newblock


\bibitem[Blasques et~al\mbox{.}(2015)]%
        {blasques2015information}
\bibfield{author}{\bibinfo{person}{Francisco Blasques}, \bibinfo{person}{Siem~Jan Koopman}, {and} \bibinfo{person}{Andre Lucas}.} \bibinfo{year}{2015}\natexlab{}.
\newblock \showarticletitle{Information-theoretic optimality of observation-driven time series models for continuous responses}.
\newblock \bibinfo{journal}{\emph{Biometrika}} \bibinfo{volume}{102}, \bibinfo{number}{2} (\bibinfo{year}{2015}), \bibinfo{pages}{325--343}.
\newblock


\bibitem[Bollerslev(1986)]%
        {bollerslev1986generalized}
\bibfield{author}{\bibinfo{person}{Tim Bollerslev}.} \bibinfo{year}{1986}\natexlab{}.
\newblock \showarticletitle{Generalized autoregressive conditional heteroskedasticity}.
\newblock \bibinfo{journal}{\emph{Journal of econometrics}} \bibinfo{volume}{31}, \bibinfo{number}{3} (\bibinfo{year}{1986}), \bibinfo{pages}{307--327}.
\newblock


\bibitem[Box et~al\mbox{.}(2015)]%
        {box2015time}
\bibfield{author}{\bibinfo{person}{George~EP Box}, \bibinfo{person}{Gwilym~M Jenkins}, \bibinfo{person}{Gregory~C Reinsel}, {and} \bibinfo{person}{Greta~M Ljung}.} \bibinfo{year}{2015}\natexlab{}.
\newblock \bibinfo{booktitle}{\emph{Time series analysis: forecasting and control}}.
\newblock \bibinfo{publisher}{John Wiley \& Sons}.
\newblock


\bibitem[Creal et~al\mbox{.}(2013)]%
        {crealGeneralizedAutoregressiveScore2013}
\bibfield{author}{\bibinfo{person}{Drew Creal}, \bibinfo{person}{Siem~Jan Koopman}, {and} \bibinfo{person}{Andr{\'e} Lucas}.} \bibinfo{year}{2013}\natexlab{}.
\newblock \showarticletitle{Generalized {{Autoregressive Score Models}} with {{Applications}}}.
\newblock \bibinfo{journal}{\emph{Journal of Applied Econometrics}} \bibinfo{volume}{28}, \bibinfo{number}{5} (\bibinfo{year}{2013}), \bibinfo{pages}{777--795}.
\newblock
\showISSN{1099-1255}
\urldef\tempurl%
\url{https://doi.org/10.1002/jae.1279}
\showDOI{\tempurl}


\bibitem[Deb et~al\mbox{.}(2017)]%
        {deb2017review}
\bibfield{author}{\bibinfo{person}{Chirag Deb}, \bibinfo{person}{Fan Zhang}, \bibinfo{person}{Junjing Yang}, \bibinfo{person}{Siew~Eang Lee}, {and} \bibinfo{person}{Kwok~Wei Shah}.} \bibinfo{year}{2017}\natexlab{}.
\newblock \showarticletitle{A review on time series forecasting techniques for building energy consumption}.
\newblock \bibinfo{journal}{\emph{Renewable and Sustainable Energy Reviews}}  \bibinfo{volume}{74} (\bibinfo{year}{2017}), \bibinfo{pages}{902--924}.
\newblock


\bibitem[Deng et~al\mbox{.}(2021)]%
        {deng2021stnorm}
\bibfield{author}{\bibinfo{person}{Jinliang Deng}, \bibinfo{person}{Xiusi Chen}, \bibinfo{person}{Renhe Jiang}, \bibinfo{person}{Xuan Song}, {and} \bibinfo{person}{Ivor~W Tsang}.} \bibinfo{year}{2021}\natexlab{}.
\newblock \showarticletitle{St-norm: Spatial and temporal normalization for multi-variate time series forecasting}. In \bibinfo{booktitle}{\emph{Proceedings of the 27th ACM SIGKDD conference on knowledge discovery \& data mining}}. \bibinfo{pages}{269--278}.
\newblock


\bibitem[Fan et~al\mbox{.}(2023)]%
        {fan2023dish}
\bibfield{author}{\bibinfo{person}{Wei Fan}, \bibinfo{person}{Pengyang Wang}, \bibinfo{person}{Dongkun Wang}, \bibinfo{person}{Dongjie Wang}, \bibinfo{person}{Yuanchun Zhou}, {and} \bibinfo{person}{Yanjie Fu}.} \bibinfo{year}{2023}\natexlab{}.
\newblock \showarticletitle{Dish-ts: a general paradigm for alleviating distribution shift in time series forecasting}. In \bibinfo{booktitle}{\emph{Proceedings of the AAAI Conference on Artificial Intelligence}}, Vol.~\bibinfo{volume}{37}. \bibinfo{pages}{7522--7529}.
\newblock


\bibitem[Godahewa et~al\mbox{.}(2021)]%
        {godahewaMonashTimeSeries2021}
\bibfield{author}{\bibinfo{person}{Rakshitha Godahewa}, \bibinfo{person}{Christoph Bergmeir}, \bibinfo{person}{Geoffrey~I. Webb}, \bibinfo{person}{Rob~J. Hyndman}, {and} \bibinfo{person}{Pablo {Montero-Manso}}.} \bibinfo{year}{2021}\natexlab{}.
\newblock \bibinfo{title}{Monash {{Time Series Forecasting Archive}}}.
\newblock
\newblock
\showeprint[arxiv]{2105.06643}~[cs, stat]


\bibitem[Guo et~al\mbox{.}(2016)]%
        {guo2016deep}
\bibfield{author}{\bibinfo{person}{Yanming Guo}, \bibinfo{person}{Yu Liu}, \bibinfo{person}{Ard Oerlemans}, \bibinfo{person}{Songyang Lao}, \bibinfo{person}{Song Wu}, {and} \bibinfo{person}{Michael~S Lew}.} \bibinfo{year}{2016}\natexlab{}.
\newblock \showarticletitle{Deep learning for visual understanding: A review}.
\newblock \bibinfo{journal}{\emph{Neurocomputing}}  \bibinfo{volume}{187} (\bibinfo{year}{2016}), \bibinfo{pages}{27--48}.
\newblock


\bibitem[Hamilton(1994)]%
        {hamilton1994ftime}
\bibfield{author}{\bibinfo{person}{JD Hamilton}.} \bibinfo{year}{1994}\natexlab{}.
\newblock \bibinfo{title}{FTime Series Analysis}.
\newblock
\newblock


\bibitem[Hornik et~al\mbox{.}(1989)]%
        {hornik1989multilayer}
\bibfield{author}{\bibinfo{person}{Kurt Hornik}, \bibinfo{person}{Maxwell Stinchcombe}, {and} \bibinfo{person}{Halbert White}.} \bibinfo{year}{1989}\natexlab{}.
\newblock \showarticletitle{Multilayer feedforward networks are universal approximators}.
\newblock \bibinfo{journal}{\emph{Neural networks}} \bibinfo{volume}{2}, \bibinfo{number}{5} (\bibinfo{year}{1989}), \bibinfo{pages}{359--366}.
\newblock


\bibitem[Hussain et~al\mbox{.}(2008)]%
        {hussain2008financial}
\bibfield{author}{\bibinfo{person}{Abir~Jaafar Hussain}, \bibinfo{person}{Adam Knowles}, \bibinfo{person}{Paulo~JG Lisboa}, {and} \bibinfo{person}{Wael El-Deredy}.} \bibinfo{year}{2008}\natexlab{}.
\newblock \showarticletitle{Financial time series prediction using polynomial pipelined neural networks}.
\newblock \bibinfo{journal}{\emph{Expert Systems with Applications}} \bibinfo{volume}{35}, \bibinfo{number}{3} (\bibinfo{year}{2008}), \bibinfo{pages}{1186--1199}.
\newblock


\bibitem[Hyndman and Koehler(2006)]%
        {hyndman2006another}
\bibfield{author}{\bibinfo{person}{Rob~J Hyndman} {and} \bibinfo{person}{Anne~B Koehler}.} \bibinfo{year}{2006}\natexlab{}.
\newblock \showarticletitle{Another look at measures of forecast accuracy}.
\newblock \bibinfo{journal}{\emph{International journal of forecasting}} \bibinfo{volume}{22}, \bibinfo{number}{4} (\bibinfo{year}{2006}), \bibinfo{pages}{679--688}.
\newblock


\bibitem[Ioffe and Szegedy(2015)]%
        {ioffe2015batch}
\bibfield{author}{\bibinfo{person}{Sergey Ioffe} {and} \bibinfo{person}{Christian Szegedy}.} \bibinfo{year}{2015}\natexlab{}.
\newblock \showarticletitle{Batch normalization: Accelerating deep network training by reducing internal covariate shift}. In \bibinfo{booktitle}{\emph{International conference on machine learning}}. pmlr, \bibinfo{pages}{448--456}.
\newblock


\bibitem[Kim et~al\mbox{.}(2021)]%
        {kimReversibleInstanceNormalization2021}
\bibfield{author}{\bibinfo{person}{Taesung Kim}, \bibinfo{person}{Jinhee Kim}, \bibinfo{person}{Yunwon Tae}, \bibinfo{person}{Cheonbok Park}, \bibinfo{person}{Jang-Ho Choi}, {and} \bibinfo{person}{Jaegul Choo}.} \bibinfo{year}{2021}\natexlab{}.
\newblock \showarticletitle{Reversible {{Instance Normalization}} for {{Accurate Time-Series Forecasting}} against {{Distribution Shift}}}. In \bibinfo{booktitle}{\emph{International {{Conference}} on {{Learning Representations}}}}.
\newblock


\bibitem[Kim et~al\mbox{.}(2004)]%
        {kim2004artificial}
\bibfield{author}{\bibinfo{person}{Tae~Yoon Kim}, \bibinfo{person}{Kyong~Joo Oh}, \bibinfo{person}{Chiho Kim}, {and} \bibinfo{person}{Jong~Doo Do}.} \bibinfo{year}{2004}\natexlab{}.
\newblock \showarticletitle{Artificial neural networks for non-stationary time series}.
\newblock \bibinfo{journal}{\emph{Neurocomputing}}  \bibinfo{volume}{61} (\bibinfo{year}{2004}), \bibinfo{pages}{439--447}.
\newblock


\bibitem[Klambauer et~al\mbox{.}(2017)]%
        {klambauer2017self}
\bibfield{author}{\bibinfo{person}{G{\"u}nter Klambauer}, \bibinfo{person}{Thomas Unterthiner}, \bibinfo{person}{Andreas Mayr}, {and} \bibinfo{person}{Sepp Hochreiter}.} \bibinfo{year}{2017}\natexlab{}.
\newblock \showarticletitle{Self-normalizing neural networks}.
\newblock \bibinfo{journal}{\emph{Advances in neural information processing systems}}  \bibinfo{volume}{30} (\bibinfo{year}{2017}).
\newblock


\bibitem[Lange et~al\mbox{.}(2022)]%
        {lange2022robust}
\bibfield{author}{\bibinfo{person}{Rutger-Jan Lange}, \bibinfo{person}{Bram van Os}, {and} \bibinfo{person}{Dick~JC van Dijk}.} \bibinfo{year}{2022}\natexlab{}.
\newblock \showarticletitle{Robust Observation-Driven Models Using Proximal-Parameter Updates}.
\newblock \bibinfo{journal}{\emph{Available at SSRN 4227958}} (\bibinfo{year}{2022}).
\newblock


\bibitem[Li et~al\mbox{.}(2018)]%
        {liAdaptiveBatchNormalization2018}
\bibfield{author}{\bibinfo{person}{Yanghao Li}, \bibinfo{person}{Naiyan Wang}, \bibinfo{person}{Jianping Shi}, \bibinfo{person}{Xiaodi Hou}, {and} \bibinfo{person}{Jiaying Liu}.} \bibinfo{year}{2018}\natexlab{}.
\newblock \showarticletitle{Adaptive {{Batch Normalization}} for Practical Domain Adaptation}.
\newblock \bibinfo{journal}{\emph{Pattern Recognition}}  \bibinfo{volume}{80} (\bibinfo{date}{Aug.} \bibinfo{year}{2018}), \bibinfo{pages}{109--117}.
\newblock
\showISSN{0031-3203}
\urldef\tempurl%
\url{https://doi.org/10.1016/j.patcog.2018.03.005}
\showDOI{\tempurl}


\bibitem[Liu et~al\mbox{.}(2024)]%
        {liu2024san}
\bibfield{author}{\bibinfo{person}{Zhiding Liu}, \bibinfo{person}{Mingyue Cheng}, \bibinfo{person}{Zhi Li}, \bibinfo{person}{Zhenya Huang}, \bibinfo{person}{Qi Liu}, \bibinfo{person}{Yanhu Xie}, {and} \bibinfo{person}{Enhong Chen}.} \bibinfo{year}{2024}\natexlab{}.
\newblock \showarticletitle{Adaptive normalization for non-stationary time series forecasting: A temporal slice perspective}.
\newblock \bibinfo{journal}{\emph{Advances in Neural Information Processing Systems}}  \bibinfo{volume}{36} (\bibinfo{year}{2024}).
\newblock


\bibitem[Lu et~al\mbox{.}(2017)]%
        {lu2017expressive}
\bibfield{author}{\bibinfo{person}{Zhou Lu}, \bibinfo{person}{Hongming Pu}, \bibinfo{person}{Feicheng Wang}, \bibinfo{person}{Zhiqiang Hu}, {and} \bibinfo{person}{Liwei Wang}.} \bibinfo{year}{2017}\natexlab{}.
\newblock \showarticletitle{The expressive power of neural networks: A view from the width}.
\newblock \bibinfo{journal}{\emph{Advances in neural information processing systems}}  \bibinfo{volume}{30} (\bibinfo{year}{2017}).
\newblock


\bibitem[Miotto et~al\mbox{.}(2018)]%
        {miotto2018deep}
\bibfield{author}{\bibinfo{person}{Riccardo Miotto}, \bibinfo{person}{Fei Wang}, \bibinfo{person}{Shuang Wang}, \bibinfo{person}{Xiaoqian Jiang}, {and} \bibinfo{person}{Joel~T Dudley}.} \bibinfo{year}{2018}\natexlab{}.
\newblock \showarticletitle{Deep learning for healthcare: review, opportunities and challenges}.
\newblock \bibinfo{journal}{\emph{Briefings in bioinformatics}} \bibinfo{volume}{19}, \bibinfo{number}{6} (\bibinfo{year}{2018}), \bibinfo{pages}{1236--1246}.
\newblock


\bibitem[Ogasawara et~al\mbox{.}(2010)]%
        {ogasawara2010adaptive}
\bibfield{author}{\bibinfo{person}{Eduardo Ogasawara}, \bibinfo{person}{Leonardo~C Martinez}, \bibinfo{person}{Daniel De~Oliveira}, \bibinfo{person}{Geraldo Zimbr{\~a}o}, \bibinfo{person}{Gisele~L Pappa}, {and} \bibinfo{person}{Marta Mattoso}.} \bibinfo{year}{2010}\natexlab{}.
\newblock \showarticletitle{Adaptive normalization: A novel data normalization approach for non-stationary time series}. In \bibinfo{booktitle}{\emph{The 2010 International Joint Conference on Neural Networks (IJCNN)}}. IEEE, \bibinfo{pages}{1--8}.
\newblock


\bibitem[Otter et~al\mbox{.}(2020)]%
        {otter2020survey}
\bibfield{author}{\bibinfo{person}{Daniel~W Otter}, \bibinfo{person}{Julian~R Medina}, {and} \bibinfo{person}{Jugal~K Kalita}.} \bibinfo{year}{2020}\natexlab{}.
\newblock \showarticletitle{A survey of the usages of deep learning for natural language processing}.
\newblock \bibinfo{journal}{\emph{IEEE transactions on neural networks and learning systems}} \bibinfo{volume}{32}, \bibinfo{number}{2} (\bibinfo{year}{2020}), \bibinfo{pages}{604--624}.
\newblock


\bibitem[Passalis et~al\mbox{.}(2019)]%
        {passalisDeepAdaptiveInput2019}
\bibfield{author}{\bibinfo{person}{Nikolaos Passalis}, \bibinfo{person}{Anastasios Tefas}, \bibinfo{person}{Juho Kanniainen}, \bibinfo{person}{Moncef Gabbouj}, {and} \bibinfo{person}{Alexandros Iosifidis}.} \bibinfo{year}{2019}\natexlab{}.
\newblock \bibinfo{title}{Deep {{Adaptive Input Normalization}} for {{Time Series Forecasting}}}.
\newblock
\newblock
\urldef\tempurl%
\url{https://doi.org/10.48550/arXiv.1902.07892}
\showDOI{\tempurl}
\showeprint[arxiv]{1902.07892}~[cs, q-fin]


\bibitem[Qui{\~n}onero-Candela et~al\mbox{.}(2022)]%
        {quinonero2022dataset}
\bibfield{author}{\bibinfo{person}{Joaquin Qui{\~n}onero-Candela}, \bibinfo{person}{Masashi Sugiyama}, \bibinfo{person}{Anton Schwaighofer}, {and} \bibinfo{person}{Neil~D Lawrence}.} \bibinfo{year}{2022}\natexlab{}.
\newblock \bibinfo{booktitle}{\emph{Dataset shift in machine learning}}.
\newblock \bibinfo{publisher}{Mit Press}.
\newblock


\bibitem[Rumelhart et~al\mbox{.}(1986)]%
        {rumelhart1986learning}
\bibfield{author}{\bibinfo{person}{David~E Rumelhart}, \bibinfo{person}{Geoffrey~E Hinton}, {and} \bibinfo{person}{Ronald~J Williams}.} \bibinfo{year}{1986}\natexlab{}.
\newblock \showarticletitle{Learning representations by back-propagating errors}.
\newblock \bibinfo{journal}{\emph{nature}} \bibinfo{volume}{323}, \bibinfo{number}{6088} (\bibinfo{year}{1986}), \bibinfo{pages}{533--536}.
\newblock


\bibitem[Salinas et~al\mbox{.}(2020a)]%
        {salinasDeepARProbabilisticForecasting2020}
\bibfield{author}{\bibinfo{person}{David Salinas}, \bibinfo{person}{Valentin Flunkert}, \bibinfo{person}{Jan Gasthaus}, {and} \bibinfo{person}{Tim Januschowski}.} \bibinfo{year}{2020}\natexlab{a}.
\newblock \showarticletitle{{{DeepAR}}: {{Probabilistic}} Forecasting with Autoregressive Recurrent Networks}.
\newblock \bibinfo{journal}{\emph{International Journal of Forecasting}} \bibinfo{volume}{36}, \bibinfo{number}{3} (\bibinfo{date}{July} \bibinfo{year}{2020}), \bibinfo{pages}{1181--1191}.
\newblock
\showISSN{0169-2070}
\urldef\tempurl%
\url{https://doi.org/10.1016/j.ijforecast.2019.07.001}
\showDOI{\tempurl}


\bibitem[Salinas et~al\mbox{.}(2020b)]%
        {salinas2020deepar}
\bibfield{author}{\bibinfo{person}{David Salinas}, \bibinfo{person}{Valentin Flunkert}, \bibinfo{person}{Jan Gasthaus}, {and} \bibinfo{person}{Tim Januschowski}.} \bibinfo{year}{2020}\natexlab{b}.
\newblock \showarticletitle{DeepAR: Probabilistic forecasting with autoregressive recurrent networks}.
\newblock \bibinfo{journal}{\emph{International journal of forecasting}} \bibinfo{volume}{36}, \bibinfo{number}{3} (\bibinfo{year}{2020}), \bibinfo{pages}{1181--1191}.
\newblock


\bibitem[Sengupta et~al\mbox{.}(2020)]%
        {sengupta2020review}
\bibfield{author}{\bibinfo{person}{Saptarshi Sengupta}, \bibinfo{person}{Sanchita Basak}, \bibinfo{person}{Pallabi Saikia}, \bibinfo{person}{Sayak Paul}, \bibinfo{person}{Vasilios Tsalavoutis}, \bibinfo{person}{Frederick Atiah}, \bibinfo{person}{Vadlamani Ravi}, {and} \bibinfo{person}{Alan Peters}.} \bibinfo{year}{2020}\natexlab{}.
\newblock \showarticletitle{A review of deep learning with special emphasis on architectures, applications and recent trends}.
\newblock \bibinfo{journal}{\emph{Knowledge-Based Systems}}  \bibinfo{volume}{194} (\bibinfo{year}{2020}), \bibinfo{pages}{105596}.
\newblock


\bibitem[Sezer et~al\mbox{.}(2020)]%
        {sezer2020financial}
\bibfield{author}{\bibinfo{person}{Omer~Berat Sezer}, \bibinfo{person}{Mehmet~Ugur Gudelek}, {and} \bibinfo{person}{Ahmet~Murat Ozbayoglu}.} \bibinfo{year}{2020}\natexlab{}.
\newblock \showarticletitle{Financial time series forecasting with deep learning: A systematic literature review: 2005--2019}.
\newblock \bibinfo{journal}{\emph{Applied soft computing}}  \bibinfo{volume}{90} (\bibinfo{year}{2020}), \bibinfo{pages}{106181}.
\newblock


\bibitem[Smyl(2020)]%
        {smylHybridMethodExponential2020}
\bibfield{author}{\bibinfo{person}{Slawek Smyl}.} \bibinfo{year}{2020}\natexlab{}.
\newblock \showarticletitle{A Hybrid Method of Exponential Smoothing and Recurrent Neural Networks for Time Series Forecasting}.
\newblock \bibinfo{journal}{\emph{International Journal of Forecasting}} \bibinfo{volume}{36}, \bibinfo{number}{1} (\bibinfo{date}{Jan.} \bibinfo{year}{2020}), \bibinfo{pages}{75--85}.
\newblock
\showISSN{0169-2070}
\urldef\tempurl%
\url{https://doi.org/10.1016/j.ijforecast.2019.03.017}
\showDOI{\tempurl}


\bibitem[Sola and Sevilla(1997)]%
        {sola1997importance}
\bibfield{author}{\bibinfo{person}{Jorge Sola} {and} \bibinfo{person}{Joaquin Sevilla}.} \bibinfo{year}{1997}\natexlab{}.
\newblock \showarticletitle{Importance of input data normalization for the application of neural networks to complex industrial problems}.
\newblock \bibinfo{journal}{\emph{IEEE Transactions on nuclear science}} \bibinfo{volume}{44}, \bibinfo{number}{3} (\bibinfo{year}{1997}), \bibinfo{pages}{1464--1468}.
\newblock


\bibitem[Ulyanov et~al\mbox{.}(2016)]%
        {ulyanov2016instance}
\bibfield{author}{\bibinfo{person}{Dmitry Ulyanov}, \bibinfo{person}{Andrea Vedaldi}, {and} \bibinfo{person}{Victor Lempitsky}.} \bibinfo{year}{2016}\natexlab{}.
\newblock \showarticletitle{Instance normalization: The missing ingredient for fast stylization}.
\newblock \bibinfo{journal}{\emph{arXiv preprint arXiv:1607.08022}} (\bibinfo{year}{2016}).
\newblock


\bibitem[van~den Oord et~al\mbox{.}(2016)]%
        {oordWaveNetGenerativeModel2016}
\bibfield{author}{\bibinfo{person}{Aaron van~den Oord}, \bibinfo{person}{Sander Dieleman}, \bibinfo{person}{Heiga Zen}, \bibinfo{person}{Karen Simonyan}, \bibinfo{person}{Oriol Vinyals}, \bibinfo{person}{Alex Graves}, \bibinfo{person}{Nal Kalchbrenner}, \bibinfo{person}{Andrew Senior}, {and} \bibinfo{person}{Koray Kavukcuoglu}.} \bibinfo{year}{2016}\natexlab{}.
\newblock \showarticletitle{{{WaveNet}}: {{A Generative Model}} for {{Raw Audio}}}.
\newblock \bibinfo{journal}{\emph{arXiv:1609.03499 [cs]}} (\bibinfo{date}{Sept.} \bibinfo{year}{2016}).
\newblock
\showeprint[arxiv]{1609.03499}~[cs]


\bibitem[Vaswani et~al\mbox{.}(2017)]%
        {vaswani2017attention}
\bibfield{author}{\bibinfo{person}{Ashish Vaswani}, \bibinfo{person}{Noam Shazeer}, \bibinfo{person}{Niki Parmar}, \bibinfo{person}{Jakob Uszkoreit}, \bibinfo{person}{Llion Jones}, \bibinfo{person}{Aidan~N Gomez}, \bibinfo{person}{{\L}ukasz Kaiser}, {and} \bibinfo{person}{Illia Polosukhin}.} \bibinfo{year}{2017}\natexlab{}.
\newblock \showarticletitle{Attention is all you need}.
\newblock \bibinfo{journal}{\emph{Advances in neural information processing systems}}  \bibinfo{volume}{30} (\bibinfo{year}{2017}).
\newblock


\bibitem[Wen et~al\mbox{.}(2017)]%
        {wen2017multi}
\bibfield{author}{\bibinfo{person}{Ruofeng Wen}, \bibinfo{person}{Kari Torkkola}, \bibinfo{person}{Balakrishnan Narayanaswamy}, {and} \bibinfo{person}{Dhruv Madeka}.} \bibinfo{year}{2017}\natexlab{}.
\newblock \showarticletitle{A multi-horizon quantile recurrent forecaster}.
\newblock \bibinfo{journal}{\emph{arXiv preprint arXiv:1711.11053}} (\bibinfo{year}{2017}).
\newblock


\bibitem[Zhang(2003)]%
        {zhangTimeSeriesForecasting2003}
\bibfield{author}{\bibinfo{person}{G.~Peter Zhang}.} \bibinfo{year}{2003}\natexlab{}.
\newblock \showarticletitle{Time Series Forecasting Using a Hybrid {{ARIMA}} and Neural Network Model}.
\newblock \bibinfo{journal}{\emph{Neurocomputing}}  \bibinfo{volume}{50} (\bibinfo{date}{Jan.} \bibinfo{year}{2003}), \bibinfo{pages}{159--175}.
\newblock
\showISSN{0925-2312}
\urldef\tempurl%
\url{https://doi.org/10.1016/S0925-2312(01)00702-0}
\showDOI{\tempurl}


\bibitem[Zhou et~al\mbox{.}(2021)]%
        {zhouInformerEfficientTransformer2021}
\bibfield{author}{\bibinfo{person}{Haoyi Zhou}, \bibinfo{person}{Shanghang Zhang}, \bibinfo{person}{Jieqi Peng}, \bibinfo{person}{Shuai Zhang}, \bibinfo{person}{Jianxin Li}, \bibinfo{person}{Hui Xiong}, {and} \bibinfo{person}{Wancai Zhang}.} \bibinfo{year}{2021}\natexlab{}.
\newblock \showarticletitle{Informer: {{Beyond Efficient Transformer}} for {{Long Sequence Time-Series Forecasting}}}.
\newblock \bibinfo{journal}{\emph{Proceedings of the AAAI Conference on Artificial Intelligence}} \bibinfo{volume}{35}, \bibinfo{number}{12} (\bibinfo{date}{May} \bibinfo{year}{2021}), \bibinfo{pages}{11106--11115}.
\newblock
\showISSN{2374-3468}
\urldef\tempurl%
\url{https://doi.org/10.1609/aaai.v35i12.17325}
\showDOI{\tempurl}


\end{thebibliography}

\end{document}